\documentclass[]{article}

\usepackage{amsmath}
\usepackage{amsfonts}
\usepackage{algpseudocode}
\usepackage{algorithm}

\usepackage{graphicx}
\usepackage[caption=true,font=normalsize,labelfont=sf,textfont=sf]{subfig} 

\usepackage{stfloats}

\title{Tensor completion using enhanced multi-mode low-rank prior and total variation}
\author{}
\date{}

\usepackage[T1]{fontenc}
\usepackage[utf8]{inputenc}
\usepackage{authblk}
\author[a]{Haijin Zeng}
\author[a]{Xiaozhen Xie \thanks{Corresponding author: xiexzh@nwafu.edu.cn}}

\author[b]{Jifeng Ning}
\affil[a]{College of Science, Northwest A\&F University, Yangling 712100, P.R. China} \affil[b]{College of Information Engineering, Northwest A\&F University, Yangling 712100, P.R. China} 

\begin{document}
\maketitle

\begin{abstract}
	
In this paper, we propose a novel model to recover a low-rank tensor by
simultaneously performing double nuclear norm regularized low-rank matrix factorizations to the all-mode matricizations of the underlying tensor. An block successive upper-bound minimization algorithm is applied to solve the model.
Subsequence convergence of our algorithm can be established, and our algorithm converges to the coordinate-wise minimizers in some mild conditions.
Several experiments on three types of public data sets show that our algorithm can recover a variety of low-rank tensors from significantly fewer samples than the other testing
tensor completion methods.

\end{abstract}




\section{Introduction}

Tensor is a generalization of vector and matrix.
A vector is a first-order or one-way tensor, and a matrix is a second-order tensor.
The results of matrix completion have been successfully applied in various practical fields,
such as inpainting \cite{Matrix_inpainting_1}, denoising \cite{Matrix_denoising_1}, image batch
alignment \cite{Matrix_image_patch_1}, key-point/saliency detection \cite{Matrix_detection_1}, and affinity learning \cite{Matrix_subspace_learning_1}.
Tensor completion as a high-order extension of matrix completion has also aroused much research interest in recent years,
due to higher-order tensor arises in many applications, for instance, video inpainting \cite{tensor_video},
magnetic resonance imaging (MRI) data recovery \cite{tensor_MRI}, 3D image reconstruction \cite{tensor_3dimage},
high-order web link analysis [16], hyperspectral or multispectral data recovery \cite{tensor_HSI},
personalized web search \cite{tensor_web}, and seismic data reconstruction \cite{tensor_seismic_data}.

Tensor completion is to recover the higher-order tensor with missing entries. Mathematically, this kind of problem can be modeled as
\begin{equation}
\label{}
\arg\min_{\mathcal{Y}} \operatorname{rank(\mathcal{Y})}, s.t.~  \mathcal{P}_{\Omega}(\mathcal{Y})=\mathcal{F},
\end{equation}		
where $\mathcal{Y} \in \mathbb{R}^{I_{1} \times \cdots \times I_{N}} $ is the underlying $N$th-order tensor;
$\mathcal{F} \in \mathbb{R}^{I_{1} \times \cdots \times I_{N}} $ is the observed data;
$\Omega$ denotes the index set of observed entries;
$\mathcal{P}_{\Omega}$ keeps the entries in $\Omega$ and zeros out others (one can find more details of $\mathcal{P}_{\Omega}$ in Section \ref{operators}).
Tensor is a high-dimensional extension of matrix,
therefore, a natural processing method is to unfold or flatten the tensor into matrix,
and then use the rank of the matrix to describe the low rank structure of the tensor, i.e.,
\begin{equation}
\label{matrix_rank}
\arg\min_{\mathbf{Y}} \operatorname{rank}(\mathbf{Y}), s.t.~  \mathcal{P}_{\Omega}(\mathcal{Y})=\mathcal{F},
\end{equation}
where $\mathbf{Y}$ is the matricization of $\mathcal{Y}$.
Unfortunately, the rank minimization in (\ref{matrix_rank}) is generally an NP-hard problem.
For effectively solving it, many methods relax the nonconvex rank function into the convex nuclear norm.
Then, the optimization problem (\ref{matrix_rank}) can be rewritten as
\begin{equation}
\label{}
\arg\min_{\mathbf{Y}} \left\|\mathbf{Y}\right\|_{*}, s.t.~  \mathcal{P}_{\Omega}(\mathcal{Y})=\mathcal{F},
\end{equation}
where $\|\cdot\|_*$ denotes the nuclear norm of a matrix.
Its solution is equivalent to the one of model (\ref{matrix_rank}) under certain conditions.
It can be solved by using algorithm such as fixed point continuation
with approximate singular value decomposition (FPCA) \cite{FPCA}, accelerated proximal gradient algorithm (APGL) \cite{APGL} or the alternating direction method \cite{ADM}.
Although the above models can recover the low-rank tensor under certain conditions,
they need convert high-dimensional tensors into 2-D matrices.
This strategy will lose useful multiorder structure information.
For instance, the spectral dimension of hyperspectral images contains imaging results of the same spatial scene in different spectral bands,
there is high correlation between the discrete spectral bands \cite{tensor_HSI};
the video often has multiple frames of images,
and there is a temporal correlation between the images of each frame \cite{wu2017robust}.

Many studies \cite{liu_tensor_2012_lrtdtv_18,yuan_tensor_2016_lrtdtv_19,cao_total_2016_lrtdtv_20,anandkumar_tensor_2016_lrtdtv_21,lu_tensor_2016_lrtdtv_22_trpca}
have proven that completion methods directly modeling tensors can better preserve the multiorder structure information than the ones modeling the tensor’ matriczation.
In the literature, two common low-rank tensor completion methods are low-rank tensor decomposition based methods and tensor rank minimization based methods, respectively.
The low-rank tensor decomposition based method generally decomposes the target tensor into a combination of several sub-tensors and matrixes for recovering a low-rank tensor from its partially observed entries,
e.g., weighted low-rank tensor decomposition method \cite{LRTD_1},
Bayes-based framework \cite{bayes_1, bayes_2},
multi-linear graph embedding \cite{muti_linear_1, Tensor_factor_prior} and tensor SVD methods \cite{tSVD_1, tSVD_2, TNN}.
These methods can effectively recover tensors, however they are usually sensitive to a given rank which is usually estimated based on the raw data.

The tensor rank minimization based method is another widely studied method, and their robustness to noisy and missing data has also been proven.
Therefore, they have been universally utilized in tensor completion problems.
Usually, they can be solved by replacing the rank function with its convex or non-convex relaxations in the minimization problem.
This type of method can significantly reduce the deviation of rank estimation.
A few notable examples are the CANDECOMP/PARAFAC rank minimization method \cite{CP},
the Tucker rank minimization method \cite{tucker_1, tucker_2},
the tensor nuclear norm (TNN) based rank approximation methods \cite{TC_TNN_1, TC_TNN_2, TC_TNN_3, TC_TNN_4, t_SVD},
and other redefined rank approximation methods with more relaxations \cite{Relaxof_Trank_1, Relaxof_Trank_2, Relaxof_Trank_3}.
Among these tensor rank minimization based methods,
the tensor singular value decomposition (t-SVD) \cite{t_SVD} based TNN,
as the tightest convex surrogate of the tensor rank,
has been widely used for low-rank tensor completion \cite{TNN}.
Specifically, the TNN regularized tensor completion model can be described as
\begin{equation}
\label{TNN}
\arg\min_{\mathcal{Y}} \left\|\mathcal{Y}\right\|_{\text{TNN}}, s.t.~  \mathcal{P}_{\Omega}(\mathcal{Y})=\mathcal{F},
\end{equation}
where $\left\|\cdot\right\|_{\text{TNN}}$ is the TNN of a tensor.
For a third-order tensor $\mathcal{A} \in \mathbb{R}^{n_{1} \times n_{2} \times n_{3}}$ and its fast Fourier transform along the third dimension $\bar{\mathcal{A}}=\operatorname{fft}(\mathcal{A},[],3)$,
the TNN of $\mathcal{A}$ is defined as the average of the nuclear norm of all the frontal slices in $\bar{\mathcal{A}}$, i.e.,
$ \|\mathcal{A}\|_{\text{TNN}} := \frac{1}{n_{3}} \sum_{i=1}^{n_{3}}\left\|\bar{\mathbf{A}}^{(i)}\right\|_{*} $, where $\bar{\mathbf{A}}^{(i)}$ denotes the $i$th frontal slice of $\bar{\mathcal{A}}$.

%

Furthermore, to alleviate bias phenomenons of the TNN minimization in tensor completion tasks,
Jiang et al. \cite{PSTNN} propose a non-convex surrogate of the tensor rank, i.e., a partial sum of the tensor nuclear norm (PSTNN).
Then PSTNN regularized tensor completion model can be written as
\begin{equation}
\label{PSTNN}
\arg\min_{\mathcal{Y}} \|\mathcal{Y}\|_{\mathrm{PSTNN}}, s.t.~  \mathcal{P}_{\Omega}(\mathcal{Y})=\mathcal{F},
\end{equation}
where $\left\|\cdot\right\|_{\text{PSTNN}}$ is the PSTNN of a tensor.
For a third-order tensor $\mathcal{A} \in \mathbb{R}^{n_{1} \times n_{2} \times n_{3}}$,
the PSTNN of $\mathcal{A}$ is defined as
$ \|\mathcal{A}\|_{\text{PSTNN}} := \frac{1}{n_{3}} \sum_{i=1}^{n_{3}}\left\|\bar{\mathbf{A}}^{(i)}\right\|_{p=M}, $
where $\|\bar{\mathbf{A}}^{(i)}\|_{p=M} := \sum_{j=M+1}^{\min (n_1, n_2)} \sigma_{j}(\bar{\mathbf{A}}^{(i)})$;
$\sigma_{j}(\bar{\mathbf{A}}^{(i)})(j=1, \cdots, \min (n_1, n_2))$ denotes the $j$-th largest singular value of $\bar{\mathbf{A}}^{(i)} \in \mathbb{C}^{n_{1} \times n_{2}}$.

\begin{figure}[!t]
\centering
\includegraphics[width=0.9\linewidth]{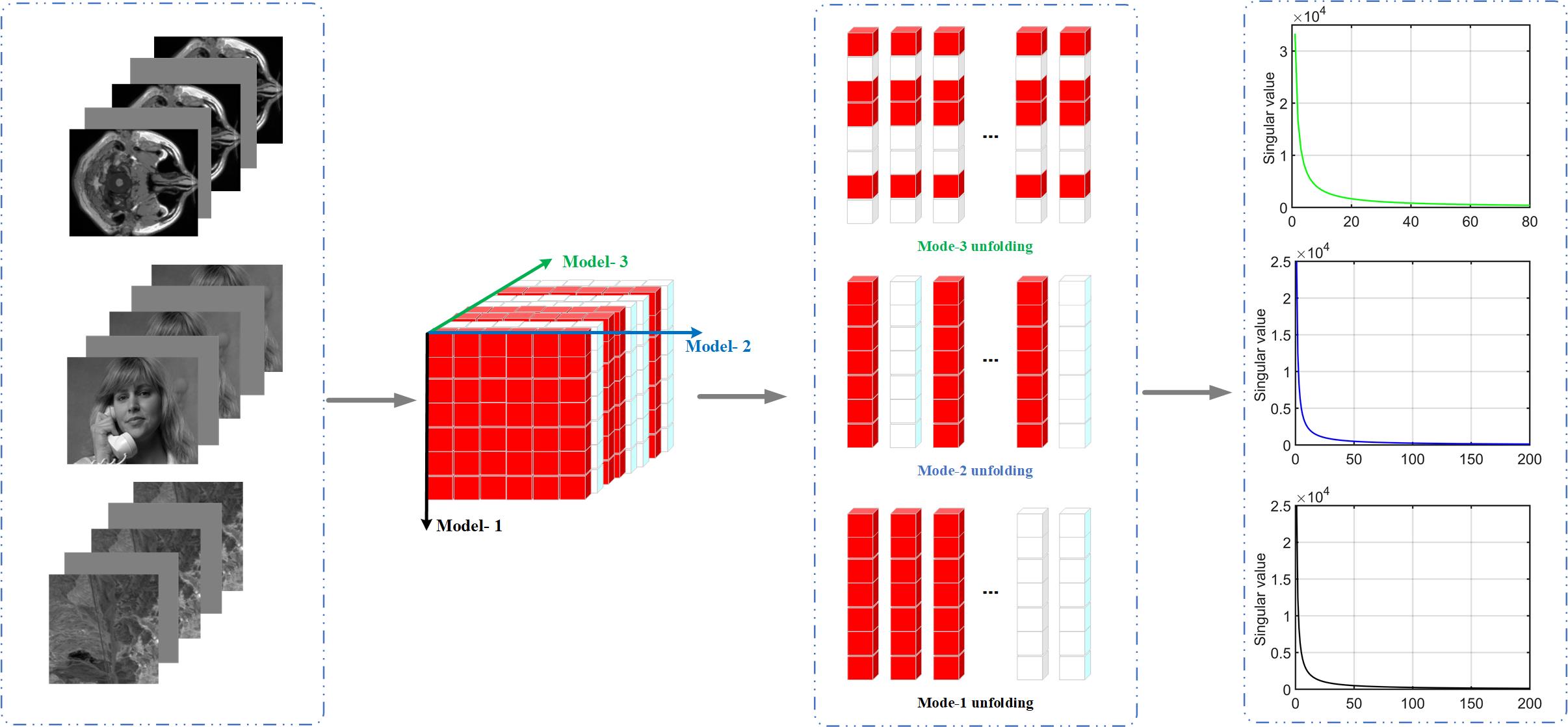}
\caption{Low rank properties of tensor mode-$n$ unfoldings}
\label{fig:modelnlowrank}
\end{figure}

Although the above-mentioned low-rank tensor completion researches show great success in dealing with various issues,
three major open questions have yet to be addressed.
Firstly, the above approaches only utilize the low-rank prior lying in one mode of the underlying tensor.
They ignore the prior knowledge of close multi-linear interactions among multiple dimensions of a given tensor object.
One can see an example in Fig. \ref{fig:modelnlowrank}.
It is obviously that all the three modes of real tensor data have similar low-rank property.
Secondly, TNN based methods \cite{TNN, PSTNN} need to compute lots of SVDs,
which become very slow or even not applicable for large-scale problems \cite{Tmac}.
Thirdly, all these methods adopt single nuclear norm or partial sum minimization of singular values norm,
which would cause suboptimal solution of the low-rank based problem.

This article presents answers to those questions.
Motivated and convinced by the much better performance of models that utilize the low-ranknesses in all mode in tensors \cite{HaLRTC, Tmac},
we could formulate a double nuclear norm based low-rank representation in all modes of underlying tensors for low-rank tensor completion tasks.
Specifically, we first apply parallel low-rank matrix factorization to each mode of the tensor.
Then, as the low-rank structure of all modes is implicitly included in the low-rank factorization,
we add the double nuclear norm regularization to the factor matrices for characterizing the underlying joint-manifold drawn from the mode factors.
By exploiting this auxiliary information, our method leverages two classic schemes and accurately estimates the mode factors and missing entries.
Then our proposed model-1 is formulated as
\begin{equation} \label{BMF_LRTC}
\begin{aligned}
&\arg\min_{\mathcal{Y},\mathbf{X}_{n}, \mathbf{A}_{n}} \sum_{n=1}^{N} (\tau_n \left\|\mathbf{X}_{n}\right\|_{\text{*}}+\lambda_n \left\|\mathbf{A}_{n}\right\|_{\text{*}}), \\
&s.t. \quad \mathcal{P}_{\Omega}(\mathcal{Y})=\mathcal{F}, \mathbf{Y}_{(n)}=\mathbf{A}_{n} \mathbf{X}_{n}, n=1,2,\cdots,N,
\end{aligned}
\end{equation}
where $\tau_{n}$ and $\lambda_n$ are positive parameters.

Further, to consider the inner geometric structure of data space,
we use the total variation (TV) regularization to construct the global relationship of real tensor data,
and propose our model-2 as follows:
\begin{equation}
\label{TVB_LRTC}
\begin{aligned}
&\arg\min_{\mathcal{Y},\mathbf{X}_{n}, \mathbf{A}_{n}} \sum_{n=1}^{N} (\tau_n \left\|\mathbf{X}_{n}\right\|_{\text{*}}+\lambda_n \left\|\mathbf{A}_{n}\right\|_{\text{*}})+\mu \left\|\mathbf{X}_{3}\right\|_{\text{TV}}, \\
& s.t. \quad \mathcal{P}_{\Omega}(\mathcal{Y})=\mathcal{F}, \mathbf{Y}_{(n)}=\mathbf{A}_{n} \mathbf{X}_{n}, n=1,2,3, \cdots, N,
\end{aligned}
\end{equation}
where $\mu$ is positive parameter; $\mathbf{A}_n$ represents a library (each column contains a signature of the $n$-th mode direction); $\mathbf{X}_n$ is called an encoding.
For example, in the unmixing problem for hyperspectral images \cite{HSI_unmixing},
each column of $\mathbf{A}_3$ denotes a spectral signature,
and each row of $\mathbf{X}_3$ denotes the fractional abundances of a given spectral signature.
This interpretation is also valid for the mode-3 factorization of videos and MRIs.
It is worth noting that the proposed model can fully capture all mode low-ranknesses and piecewise smooth prior of the underlying tensor,
and thus is expected to have a strong ability of low-rank tensor completion.
For the other details of the models, we ask for the readers patience until Section \ref{Section:proposed model}.

\section{Preliminary}

Before introducing our models and their algorithms,
we review some notations, tensor operations, regularizers with physical meaning and operators.

\subsection{Notations}

Following \cite{Tmac}, vectors are denoted as bold lower-case letters, e.g., $\mathbf{x}, \mathbf{y}$;
matrices are denoted as bold upper-case letters, e.g., $\mathbf{X}, \mathbf{Y}$;
and tensors are denoted as caligraphic letters, e.g., $\mathcal{X}, \mathcal{Y}$.
Let $x_{i_{1}, \cdots, i_{N}}$ represents the $\left(i_{1}, \cdots, i_{N}\right)$th component of an $N$th-order tensor $\mathcal{X}$.
Then, for $\mathcal{X}, \mathcal{Y} \in \mathbb{R}^{I_{1} \times \cdots \times I_{N}}$, their inner product is defined as
\begin{equation}
\langle\mathcal{X}, \mathcal{Y}\rangle=\sum_{i_{1}=1}^{I_{1}} \cdots \sum_{i_{N}=1}^{I_{N}} x_{i_{1}, \cdots, i_{N}} y_{i_{1}, \cdots, i_{N}}.
\end{equation}
Based on the inner product, one can define the  \textbf{Frobenius norm} of a tensor $\mathcal{X}$ as $\|\mathcal{X}\|_{\text{F}}=\sqrt{\langle\mathcal{X}, \mathcal{X}\rangle}$.
\textbf{Fiber} of tensor $\mathcal{X}$ are defined as a vector obtained by fixing all indices of $\mathcal{X}$ except one,
and \textbf{Slice} of $\mathcal{X}$ are defined as a matrix by fixing all indices of $\mathcal{X}$ except two.
The \textbf{mode-$n$ matricization}/\textbf{unfolding} of $\mathcal{X}$ is denoted as a matrix $\mathbf{X}_{(n)} \in \mathbb{R}^{I_{n} \times \Pi_{j \neq n} I_{j}}$
with columns being the mode-$n$ fibers of $\mathcal{X}$ in the lexicographical order.

To clearly represent the matricization process, we define $\operatorname{unfold}_{n}(\mathcal{X})=\mathbf{X}_{(n)}$,
and $\operatorname{fold}_{n}$ is the inverse of $\operatorname{unfold}_{n}$, i.e.,
$\operatorname{fold}_{n}\left(\operatorname{unfold}_{n}(\mathcal{X})\right)=\mathcal{X}$.
Let $\operatorname{rank}_{n}(\mathcal{X}) = \operatorname{rank}(\mathbf{X}_{(n)})$ denote the $n$-rank of $\mathcal{X}$.
Then the rank of $\mathcal{X}$ is defined as an array $\operatorname{rank}(\mathcal{X})=\left(\operatorname{rank}\left(\mathbf{X}_{(1)}\right), \ldots, \operatorname{rank}\left(\mathbf{X}_{(N)}\right)\right)$.

\subsection{Total variation}
In (\ref{TVB_LRTC}), $\mathbf{X}_{3}^{(i, k)} \in \mathbb{R}^{1 \times I_{1} I_{2}}$ denotes a vector by  lexicographically ordering the entries of the matrix
$\mathcal{X}\left(:,:, k, i_{4}, \ldots, i_{N}\right) \in$ $\mathbb{R}^{I_{1} \times I_{2}}$ which is a slice of the tensor $\mathcal{X}$,
where $i=1+\sum_{p=4}^{N}\left(i_{p}-1\right) J_{p}$ and $J_{p}=\Pi_{m=4}^{p-1} I_{m}$;
$$\mathcal{X}= \operatorname{fold} _{3}\left(\mathbf{X}_{3}\right) \in \mathbb{R}^{I_{1} \times I_{2} \times r_{3} \times I_{4} \times \cdots \times I_{N}}.$$
Then, the isotropic \textbf{total variation (TV)} is defined as follows:
\begin{equation}
\label{equation_TV}
\|\mathbf{X}_{3}\|_{\text{TV}}:=\sum_{k=1}^{S} \sum_{i=1}^{r_{n}} \sum_{j=1}^{n_{2}} \sqrt{\left|\tilde{\mathbf{D}}_{j, 1} \mathbf{X}_{3}^{(i, k)}\right|^{2}+\left|\tilde{\mathbf{D}}_{j, 2} \mathbf{X}_{3}^{(i, k)}\right|^{2}}
\end{equation}
where $\mathbf{X}_{3}^{(i, k)}$ represents the $k$-th block of  $i$-th row of $\mathbf{X}_{3}$,
$\tilde{\mathbf{D}}_{j, 1}$ and $\tilde{\mathbf{D}}_{j, 2}$ represent the discrete gradient operators at the $1 \mathrm{st}$- and $2 \mathrm{nd}$- mode directions, respectively.
Following the representation of $\tilde{\mathbf{D}}$, $\tilde{\mathbf{D}}_{j, 1} \mathbf{X}_{3}^{(i, k)}$ denotes the gradient values of $\mathbf{X}_{3}^{(i, k)}$ at the 1st mode directions and $\tilde{\mathbf{D}}_{j, 2} \mathbf{X}_{3}^{(i, k)}$ denotes the gradient values of $\mathbf{X}_{3}^{(i, k)}$ at the 2nd mode directions of the $j$th pixel in $\mathbf{X}_{3}^{(i, k)}.$

\subsection{Operators} \label{operators}
The \textbf{Proximal Operator} of a given convex function $f(x)$ is defined as
\begin{equation}
\label{PPA}
\operatorname{prox}_{f}(x,y):=\arg \min _{x} f(x)+\frac{\rho}{2}\|x-y\|^{2},
\end{equation}
where $\rho$ is a positive constant.
Friendly, the problem $\arg\min _{x}\{f(x)\}$ is equivalent to $\arg\min _{x, y}\left\{f(x)+\frac{\rho}{2}\|x-y\|^{2}\right\}$.
Thus one can obtain the minimization of $f(x)$ by iteratively solving prox$_{f}\left(x, x^{k}\right)$, where $x^{k}$ is the latest update of $x$.
The highlight of the proximal operator is that it can guarantee the strong convexity of objective function (\ref{PPA}), as long as $f(x)$ is convex.

Let $\Omega$ be the index set of observed entries, then the \textbf{Projection operator} $\mathcal{P}_{\Omega}$ keeps the entries in $\Omega$ and zeros out others, i.e.,
\begin{equation}
\left(\mathcal{P}_{\Omega}(\mathcal{Y})\right)_{i_{1} \cdots i_{N}}=\left
\{\begin{array}{ll}
{y_{i_{1}, \cdots, i_{N}},} & {\left(i_{1}, \cdots, i_{N}\right) \in \Omega} \\
{0,} & {\text { otherwise }}
\end{array}\right.
\end{equation}

The \textbf{singular value shrinkage (SVT) operator} \cite{SVT} is defined as follows.
Supposing $\mathbf{M}$ is a matrix of size $I_1I_2 \times I_3$, and the singular value of matrix $\mathbf{M}$ of rank $r$ is decomposed into
\begin{equation*}
\mathbf{M}=\mathbf{P}\mathbf{E}_{r} \mathbf{Q}^{*}, \mathbf{E}_{r}=\operatorname{diag}\left(\left\{\sigma_{i}\right\}_{1 \leq i \leq r}\right)
\end{equation*}
The singular value shrinkage operator then obeys
\begin{equation*}
\operatorname{SH}_{\delta}(\mathbf{M})=\arg \min_{\operatorname{rank}(\mathbf{X}) \leq r} \delta\|\mathbf{X}\|_{*}+\frac{1}{2}\|\mathbf{X}-\mathbf{W}\|_{\text{F}}^{2},
\end{equation*}
where
\begin{equation}
\label{SVT}
\operatorname{SH}_{\delta}(\mathbf{W})=\mathbf{P} \operatorname{diag}\left\{\max \left(\left(\sigma_{i}-\delta\right), 0\right)\right\} \mathbf{Q}^{*}.
\end{equation}

\section{Proposed models and algorithms}
\label{Section:proposed model}

\subsection{Proposed models}
The objective function of our model-1 (\ref{BMF_LRTC}) is as following:
\begin{equation}
f(\mathbf{X}, \mathbf{A}, \mathcal{Y})= \sum_{n=1}^{N} (\frac{\alpha_{n}}{2}\left\|\mathbf{Y}_{(n)}-\mathbf{A}_{n} \mathbf{X}_{n}\right\|_{\text{F}}^{2}+\tau_n \left\|\mathbf{X}_{n}\right\|_{\text{*}}+\lambda_n \left\|\mathbf{A}_{n}\right\|_{\text{*}}),
\end{equation}
where $\alpha_n, n=1,2,\cdots,N$, are positive weights satisfying $\sum_{n=1}^{N}\alpha_{n}=1$.

The objective function of our model-2 (\ref{TVB_LRTC}) is as following:
\begin{equation} \label{model2}
f(\mathbf{X}, \mathbf{A}, \mathcal{Y})= \sum_{n=1}^{N} (\frac{\alpha_{n}}{2}\left\|\mathbf{Y}_{(n)}-\mathbf{A}_{n} \mathbf{X}_{n}\right\|_{\text{F}}^{2}+\tau_n \left\|\mathbf{X}_{n}\right\|_{\text{*}}+\lambda_n \left\|\mathbf{A}_{n}\right\|_{\text{*}})+\mu \left\|\mathbf{X}_{3}\right\|_{\text{TV}}.
\end{equation}

Firstly, we explain the reason why we constrain the low-rank property in all modes of underlying tensors.
In an $n$th-order tensor, each order represents one factor and has its specific inherent structural properties.
Therefore, each mode of the underlying tensor has specific prior information.
Although a tensor could be comprised of randomly arranged elements,
it is usually assumed that the within-factor and joint-factor variations are known a priori and can be regarded as auxiliary information \cite{Tensor_factor_prior}.
For example, a video object is a third-order tensor with variations spanned by the rows, columns, and time axis.
Even when the value of an element is unknown, we may reasonably infer that adjacent rows, columns or frames are highly correlated.
This is because the local similarity of visual data usually exists in within-factor relations (e.g., between adjacent rows, columns or frames)
or joint-factor relations (e.g., between spatially adjacent and temporally adjacent pixels).
See Fig. \ref{fig:modelnlowrank} for an illustration of real 3rd-order tensor data.
It is obviously seen that the singular value curves of their three modes decay rapidly,
that is to say that only a small part of the singular values are greater than zero.
Therefore, the three modes of the real tensor data have the similar low-rank property.
Actually, this phenomenon has specific physical meaning.
Take the hyperspectral image (HSI) $\mathcal{Y} \in \mathbb{R}^{m \times n \times p}$ for an example,
it is well known that each spectral characteristic can be represented by a linear combination of a small number of pure spectral endmembers.
It means that its mode-3 matricization $\mathbf{Y}_{(3)}$ can be decomposed into $\mathbf{Y}_{(3)}=\mathbf{A}_3\mathbf{X}_3$,
where $\mathbf{A}_3 \in \mathbb{R}^{p \times r}$ is the so-called endmember matrix,
and $\mathbf{X}_3 \in \mathbb{R}^{r \times mn}$ is regarded as the abundance matrix.
As described in \cite{LRTDTV}, the number of endmembers $r$ is relatively small, i.e., $r \ll p$ or $r \ll m n$.
That is to say that only a small part of the singular values are greater than zero, as shown in the fourth column of Fig. \ref{fig:modelnlowrank}.
Based on the above practical physical meaning,
we utilize the low-rank prior lying in all modes of underlying tensors to promote the performance of tensor completion models.

Secondly, we explain the reason why we adopt the double nuclear norms of $\mathbf{A}_n$ and $\mathbf{X}_n$ to represent the low-rank prior in each mode.
Without increasing the computational complexity, instead of the traditional single decomposition,
each mode of the tensor is decomposed into two smaller factor matrices \cite{Tmac}, i.e., $\mathbf{Y}_{(n)}=\mathbf{A}_{n}\mathbf{X}_{n}, n=1,2,\cdots,N$.
The low-rank structure of tensors not only is inherited by the factor matrices, i.e., $\mathbf{A}_{n}$, $\mathbf{X}_{n}$, but also can be represented more sufficiently.
Then, we add the double nuclear norm regularization to the factor matrices for characterizing the underlying joint-manifold drawn from the mode factors.
By exploiting this auxiliary information, our method leverages two classic schemes and accurately estimates the model factors and missing entries.
Unfortunately, it is difficult to directly calculate the nuclear norm of the product of two matrices,
i.e., $\|\mathbf{Y}_{(n)}\|_{*}=\|\mathbf{A}_{n}\|_{*}\|\mathbf{X}_{n}\|_{*}, n=1,2,3, \cdots, N$.
Therefore, according to the fundamental inequality, we reformulate the product of two nuclear norm into
\begin{equation}
\begin{aligned}
\arg\min_{\mathbf{Y}_{(n)}} \|\mathbf{Y}_{(n)}\|_{*} &=\arg\min _{ \mathbf{Y}_{(n)}=\mathbf{A}_n \mathbf{X}_n} \|\mathbf{A}_n\|_{*}\|\mathbf{X}_n\|_{*} \\
&=\arg\min _{\mathbf{Y}_{(n)}=\mathbf{A}_n \mathbf{X}_n}\left(\frac{1}{2}\left(\|\mathbf{A}_n\|_{*}+\|\mathbf{X}_n\|_{*}\right)\right)^{2}.
\end{aligned}
\end{equation}

\begin{figure}
	\centering
	\includegraphics[width=0.9\linewidth]{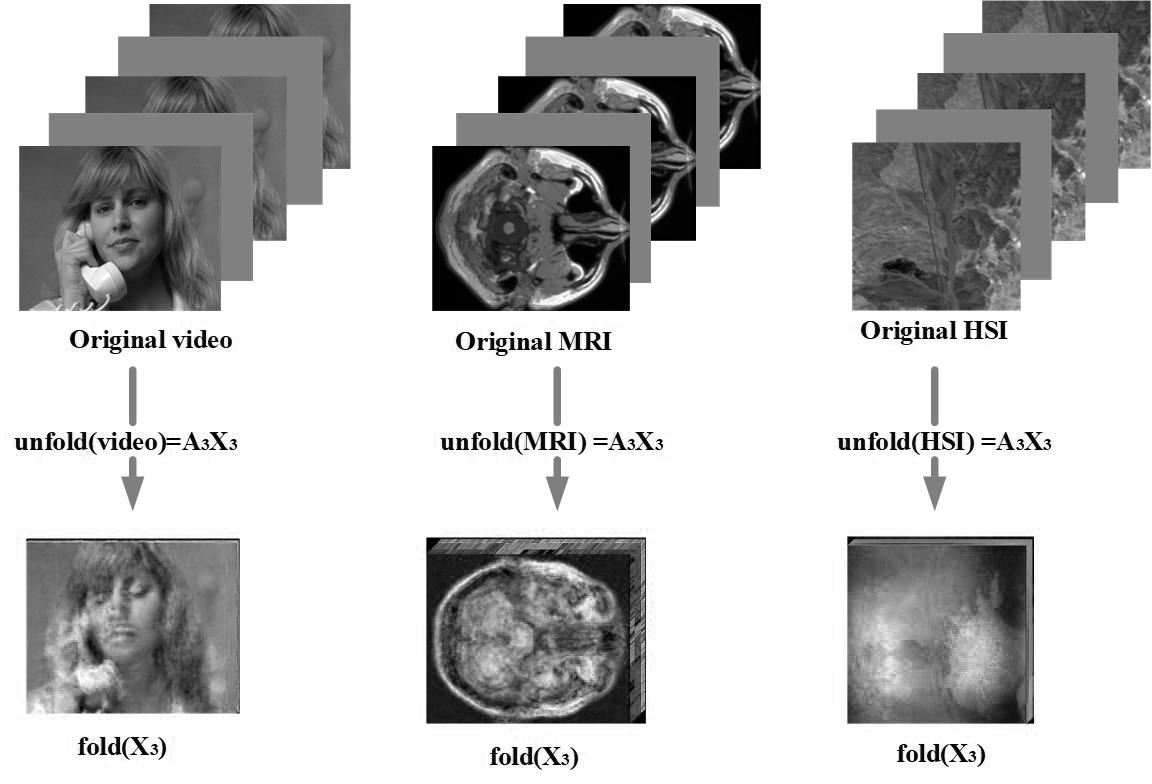}
	\caption{Illustration of the structural characteristics of $\mathbf{X_3}$. }
	\label{fig:x3flowchat}
\end{figure}

Thirdly, we explain why we introduce the TV regularization of $\mathbf{X}_3$ to the proposed low-rank tensor completion model.
The TV regularization measures the difference between a pixel and its neighbors.
The smaller the difference is, the better the TV regularization plays.
Because the data is piecewise smooth with respect to the 1st- and 2nd-mode direction,
the difference between the pixel of $\mathbf{X}_n$ and its 1st- and 2nd-mode direction neighbors is small.
Thus, we can introduce the TV regularization of $\mathbf{X}_{n}$ at the 1st-and 2nd-mode direction into the tensor completion problem.
However, $\mathbf{X}_{1}$ and $\mathbf{X}_{2}$ do not contain the complete information of the 1st- and 2nd mode for $\mathcal{Y}$,
because the rank of $\mathbf{Y}_{(1)}$ is $r_{1}\left(r_{1}<I_{1}\right)$,
that is, the dimension of the corresponding tensor is $r_{1} \times I_{2} \times \cdots \times I_{N}$ \cite{MFTV}.
Thus, we introduce the TV regularization of $\mathbf{X}_{n_{0}}, n_{0} \in\{3,4, \ldots, N\} .$ Without loss of generality, we adopt the TV regularization of $\mathbf{X}_{3}$.
For three types of public tensor datasets, we show the specific structure of their $\mathbf{X}_{3}$ in Fig. \ref{fig:x3flowchat}. As shown in Fig. \ref{fig:x3flowchat}, $\mathbf{X}_{3}$ has an obvious smooth structure, so it is appropriate to use TV to explore the inherent structure prior of $\mathbf{X}_{3}$.

\subsection{Proposed algorithms}

The proposed model-1 (\ref{BMF_LRTC}) and model-2 (\ref{TVB_LRTC}) are two complicated optimization problems, which are difficult to solve directly.
Here, we adopt the block successive upper-bound minimization (BSUM)\cite{BSUM} to solve them.

According to the proximal operator (\ref{PPA}), the update can be written as:
\begin{equation}
\label{equation_original_PPA}
\operatorname{Prox}_{f}(\mathcal{S}, \mathcal{S}^k)= \arg\min_{\mathcal{S}} f\left(\mathcal{S}\right)+\frac{\rho}{2}\left\|\mathcal{S}-\mathcal{S}^k\right\|_{\text{F}}^{2},
\end{equation}
where $\rho>0$ is the proximal parameter, $\mathcal{S}=(\mathbf{X}, \mathbf{A}, \mathcal{Y})$ and $\mathcal{S}^{k}=\left(\mathbf{X}^{k}, \mathbf{A}^{k}, \mathcal{Y}^{k}\right)$.

Let $S_{1}^{k}=\left(\mathbf{X}^{k}, \mathbf{A}^{k}, \mathcal{Y}^{k}\right)$,
$S_{2}^{k}=\left(\mathbf{X}^{k+1}, \mathbf{A}^{k}, \mathcal{Y}^{k}\right)$,
$S_{3}^{k}=\left(\mathbf{X}^{k+1}, \mathbf{A}^{k+1}, \mathcal{Y}^{k}\right)$.
By BSUM, (\ref{equation_original_PPA}) can be rewritten as follows:
\begin{equation}
\label{equation:XAY}
\left\{\begin{array}{l}
\displaystyle \mathbf{X}^{k+1}=  \operatorname{Prox}_{f}\left(\mathbf{X}, \mathcal{S}_{1}^{k}\right)= \arg\min_{\mathbf{X}} f\left(\mathbf{X}, \mathbf{A}^{k}, \mathcal{Y}^{k}\right)+\frac{\rho}{2}\left\|\mathbf{X}-\mathbf{X}^{k}\right\|_{\text{F}}^{2}, \\
\displaystyle\mathbf{A}^{k+1}=  \operatorname{Prox}_{f}\left(\mathbf{A}, \mathcal{S}_{2}^{k}\right)= \arg\min_{\mathbf{A}} f\left(\mathbf{X}^{k+1}, \mathbf{A}, \mathcal{Y}^{k}\right)+\frac{\rho}{2}\left\|\mathbf{A}-\mathbf{A}^{k}\right\|_{\text{F}}^{2}, \\
\displaystyle\mathcal{Y}^{k+1}= \operatorname{Prox}_{f}\left(\mathcal{Y}, \mathcal{S}_{3}^{k}\right)= \arg\min_{\mathcal{Y}} f\left(\mathbf{X}^{k+1}, \mathbf{A}^{k+1}, \mathcal{Y}\right)+\frac{\rho}{2}\left\|\mathcal{Y}-\mathcal{Y}^{k}\right\|_{\text{F}}^{2}.
\end{array}\right.	
\end{equation}

\subsubsection{Update $\mathbf{X}_n$ with fixing others}

The $\mathbf{X}_n$-sub-problem in (\ref{equation:XAY}) can be written as follows:
\begin{equation}
\label{equation_X}
\mathbf{X}_n^{k+1}=\arg\min_{\mathbf{X}_n} \sum_{n=1}^{N} (\frac{\alpha_{n}}{2}\left\|\mathbf{Y}_{(n)}-\mathbf{A}_{n} \mathbf{X}_{n}\right\|_{\text{F}}^{2}+\tau_n \left\|\mathbf{X}_{n}\right\|_{\text{*}}+\frac{\rho_n}{2}\left\|\mathbf{X}_n-\mathbf{X}_n^{k}\right\|_{\text{F}}^{2})+\mu \left\|\mathbf{X}_{3}\right\|_{\text{TV}}.
\end{equation}
To efficiently solve it,
we first introduce one auxiliary variable.
Then (\ref{equation_X}) can be rewritten as
\begin{equation}
\label{equation_X_aux}
\begin{aligned}
\arg\min_{\mathbf{X}_n, \mathbf{Z}_n}& \sum_{n=1}^{N} (\frac{\alpha_{n}}{2}\left\|\mathbf{Y}_{(n)}-\mathbf{A}_{n} \mathbf{X}_{n}\right\|_{\text{F}}^{2}+\tau_n \left\|\mathbf{Z}_{n}\right\|_{\text{*}}
+\frac{\rho_n}{2}\left\|\mathbf{X}_n-\mathbf{X}_n^{k}\right\|_{\text{F}}^{2})
+\mu \left\|\mathbf{X}_{3}\right\|_{\text{TV}}, \\
& s.t., \mathbf{X}_{n}=\mathbf{Z}_{n}.
\end{aligned}
\end{equation}
Based on the augmented Lagrange multiplier (ALM) method, the above minimization problem (\ref{equation_X_aux}) can be transformed into
\begin{equation}
\label{equation:X_alm}
\begin{aligned}
\arg\min_{\mathbf{X}_n, \mathbf{Z}_n} & \sum_{n=1}^{N} (\frac{\alpha_{n}}{2}\left\|\mathbf{Y}_{(n)}-\mathbf{A}_{n} \mathbf{X}_{n}\right\|_{\text{F}}^{2}+\tau_n \left\|\mathbf{Z}_{n}\right\|_{\text{*}}
+\frac{\rho_n}{2}\left\|\mathbf{X}_n-\mathbf{X}_n^{k}\right\|_{\text{F}}^{2}\\
&+\left\langle\Gamma_{n}^\mathbf{X}, \mathbf{X}_n-\mathbf{Z}_n\right\rangle+\frac{\rho_{n}}{2}\left\|\mathbf{X}_{n}-\mathbf{Z}_{n}\right\|_{\text{F}}^{2})+\mu \left\|\mathbf{X}_{3}\right\|_{\text{TV}},
\end{aligned}
\end{equation}
where $\Gamma_{n}^\mathbf{X}$ is a Lagrange multiplier.
With other variables fixed, the minimization subproblem for $\mathbf{Z}_n$ can be deduced from (\ref{equation:X_alm}) as follows:
\begin{equation}
\displaystyle \mathbf{Z}_n^{k+1}= \arg\min_{\mathbf{Z}_n} \tau_n \left\|\mathbf{Z}_{n}\right\|_{\text{*}}+\frac{\rho_{n}}{2}\left\|\mathbf{X}_{n}^{k}-\mathbf{Z}_{n}+\Gamma_{n}^{\mathbf{X}}/\rho_n\right\|_{\text{F}}^{2}.
\end{equation}
By using the SVT operator (\ref{SVT}), it is easy to get
\begin{equation}
\label{equation_solution_Zn}
\begin{aligned}
\mathbf{Z}_n^{k+1}=\operatorname{SH}_{\frac{\tau_{n}}{\rho_n}}(\mathbf{X}_{n}^{k}+\Gamma_{n}^{\mathbf{X}}/\rho_n), n=1,2, \cdots, N.
\end{aligned}
\end{equation}

Based on the ALM method, the multipliers are updated by the following equations:
\begin{equation}
\label{equation:Lambda_0}
\Gamma_{n}^{\mathbf{X}} = \Gamma_{n}^{\mathbf{X}} + \mathbf{X}_n-\mathbf{Z}_n.
\end{equation}

With other variables fixed, the minimization subproblem for $\mathbf{X}_n (n\not=3)$ can be deduced from (\ref{equation:X_alm}) as follows:
\begin{equation}
\label{equationforX}
\begin{aligned}
\mathbf{X}_n^{k+1}&= \arg\min_{\mathbf{X}_n}\frac{\alpha_{n}}{2}\left\|\mathbf{Y}_{(n)}-\mathbf{A}_{n}^{k} \mathbf{X}_{n}\right\|_{\text{F}}^{2}+\frac{\rho_{n}}{2}\left\|\mathbf{X}_{n}-\frac{\mathbf{Z}_{n}^{k+1}-\Gamma_{n}^k/\mu_n+\mathbf{X}_n^{k}}{2}\right\|_{\text{F}}^{2}.
\end{aligned}
\end{equation}
They are convex and have the following closed-form solutions
\begin{equation}
\label{equation:solution_Xn}
\begin{aligned}
\mathbf{X}_n^{k+1}=(\alpha_{n}\mathbf{A}_n^T\mathbf{A}_n+2\rho \mathbf{I}_n)^{-1}[\alpha_{n}\mathbf{A}_n^T\mathbf{Y}_{(n)}+\mu_n (\frac{\mathbf{Z}_{n}^{k+1}-\Gamma_{n}^k/\mu_n+\mathbf{X}_n^{k}}{2}).
\end{aligned}
\end{equation}

With other variables fixed, the minimization subproblem for $\mathbf{X}_{3}$ can be deduced from (\ref{equation:X_alm}) as follows:
\begin{equation}
\label{equation_X3_ori}
\begin{aligned}
\mathbf{X}_3^{k+1}&= \arg\min_{\mathbf{X}_3}\frac{\alpha_{3}}{2}\left\|\mathbf{Y}_{(3)}-\mathbf{A}_{3}^{k} \mathbf{X}_{3}\right\|_{\text{F}}^{2}+\frac{\rho_{3}}{2}\left\|\mathbf{X}_{3}-\frac{\mathbf{Z}_{3}^{k+1}-\Gamma_{3}^k/\mu_3+\mathbf{X}_3^{k}}{2}\right\|_{\text{F}}^{2}+\mu \left\|\mathbf{X}_{3}\right\|_{\text{TV}}.
\end{aligned}
\end{equation}
Compared with the optimization problem of $X_n, n\not=3$,
the optimization problem of $\mathbf{X}_3$ has an additional TV regular term imposed on $\mathbf{X}_3$.
It can be solved efficiently using ADMM \cite{ADM3D, ADMYin, MFTV}.
To obtain the closed solution of (\ref{equation_X3_ori}),
we denote $\hat{\mathbf{X}}$ as the transpose of $\mathbf{X}$.
Then, the solution of (\ref{equation_X3_ori}) is equivalent to one of the following minimization problem:
\begin{equation}
\label{equation_X3_T}
\hat{\mathbf{X}}_3^{k+1} = \arg\min_{\hat{\mathbf{X}}_{3}} \frac{1}{2}\left\|\hat{\mathbf{Y}}_{(3)}^{k}-\hat{\mathbf{X}}_{3} \hat{\mathbf{A}}_{3}^{k}\right\|_{\text{F}}^{2}+\frac{\rho}{2}\left\|\hat{\mathbf{X}}_{3}-\frac{\hat{\mathbf{Z}}_{3}^{k+1}-\hat{\Gamma}_{3}^k/\mu_3+\hat{\mathbf{X}}_3^{k}}{2}\right\|_{\text{F}}^{2}+\mu \left\|\hat{\mathbf{X}}_{3}\right\|_{\text{TV}},
\end{equation}
where $\rho= \frac{\rho_3}{\alpha_{3}}$.
For simplicity, let $\hat{\mathbf{O}}_3^k=\frac{\hat{\mathbf{Z}}_{3}^{k+1}-\hat{\Gamma}_{3}^k/\mu_3+\hat{\mathbf{X}}_3^{k}}{2}$.
Then, we introduce two auxiliary variables and convert (\ref{equation_X3_T}) into
\begin{equation}
\label{equation_X3}
\begin{aligned}
&\arg\min _{\hat{\mathbf{X}}_{3}, \mathbf{U}} \mu \sum_{i=1}^{s_{3}} \sum_{j=1}^{r_{3}}\left\|\mathbf{U}_{i, j}\right\|_{2}+\frac{1}{2}\left\|\tilde{\mathbf{Y}}_{(3)}^{k}-\hat{\mathbf{X}}_{3} \hat{\mathbf{A}}_{3}^{k}\right\|_{\text{F}}^{2}+\frac{\rho}{2}\left\|\hat{\mathbf{X}}_{3}-\hat{\mathbf{O}}_{3}^{k}\right\|_{\text{F}}^{2} \\
&s.t. \quad \mathbf{U}_{1}=\mathbf{D}_{1} \hat{\mathbf{X}}_{3}, \mathbf{U}_{2}=\mathbf{D}_{2} \hat{\mathbf{X}}_{3}
\end{aligned}
\end{equation}
where $\mathbf{U}_{i, j}=\left[\left(\mathbf{U}_{1}\right)_{i, j},\left(\mathbf{U}_{2}\right)_{i, j}\right] \in \mathbb{R}^{1 \times 2}$,
$\left(\mathbf{U}_{1}\right)_{i, j}$ and $\left(\mathbf{U}_{2}\right)_{i, j}$ denote the $(i, j)$th entries of $\mathbf{U}_{1}$ and $\mathbf{U}_{2},$ respectively;
$\mathbf{D}_{t}:=\operatorname{Diag}(\tilde{{\mathbf{D}}}_{t}, \tilde{{\mathbf{D}}}_{t}, \cdots, \tilde{{\mathbf{D}}}_{t})$,
$t=1,2$,
and $\tilde{\mathbf{D}}_{1}$ and $\tilde{\mathbf{D}}_{2}$ are respectively the assembled first-order difference matrices in the 1st- and 2nd-mode directions
based on $\tilde{\mathbf{D}}_{j, 1}$ and $\tilde{\mathbf{D}}_{j, 2}$ in ( \ref{equation_TV}).

The problem (\ref{equation_X3}) can be solved by solving two decoupled subproblems, which the convergence can be guaranteed \cite{glowinski2008lectures}.
By ALM method, (\ref{equation_X3}) can be rewritten as
\begin{equation}
\label{equation_alm_X3}
\begin{aligned}
\arg\min_{\hat{\mathbf{X}}_{3},\mathbf{U}} & \frac{1}{2}\left\|\hat{\mathbf{Y}}_{(3)}^{k}-\hat{\mathbf{X}}_{3} \hat{\mathbf{A}}_{3}^{k}\right\|_{\text{F}}^{2}+ \frac{\rho}{2}\left\|\hat{\mathbf{X}}_{3}-\hat{\mathbf{O}}_{3}^{k}\right\|_{\text{F}}^{2}+\mu \sum_{i=1}^{\mathrm{s_3}} \sum_{j=1}^{\mathrm{r_3}}\left\|\mathbf{U}_{i,j}\right\|_{2}\\
&+\left\langle\Lambda, \mathbf{B} \hat{\mathbf{X}}_{3}+\mathrm{\mathbf{C}} \mathbf{U}\right\rangle+ \frac{\beta}{2}\left\|\beta \hat{\mathbf{X}}_{3}+\mathrm{\mathbf{C}\mathbf{U}}\right\|_{\text{F}}^{2},
\end{aligned}
\end{equation}
where
$\mathbf{B} \hat{\mathbf{X}}_{3}+\mathbf{C} \mathbf{U}:=\left[{\mathbf{D}_{1}},{\mathbf{D}_{2}}\right]^T \hat{\mathbf{X}}_{3}-\hat{\mathbf{I}}_{2s_3 \times 2s_3} \left[{\mathbf{U}_{1}}, {\mathbf{U}_{2}}\right]^T=\mathbf{0}_{2s_3 \times r_3}$ is a convenient form of the constraints in (\ref{equation_X3}), and $\hat{\mathbf{I}}_{i \times i}$ is the $i$-by-$i$ identity matrix;
$\Lambda= \left(\Lambda_{1}, \Lambda_{2}\right)^{T}$; $\beta>0$ is the penalty parameter.
Then, (\ref{equation_alm_X3}) can be solved by alternately iterating the three variables $\hat{\mathbf{X}}_{3}, \mathbf{U}$ and $\Lambda$.
Specifically, let $p$ denotes the iteration indicator for solving the problem (\ref{equation_X3}).
With other variables fixed, for the $\mathbf{X}_{3}$-subproblem, we have
\begin{equation}
\label{equation_X3_subsub}
\begin{aligned}
\hat{\mathbf{X}}_{3}^{k+1, p+1} = \arg\min_{\hat{\mathbf{X}}_{3}} & \frac{1}{2}\left\|\hat{\mathbf{Y}}_{(3)}^{k}-\hat{\mathbf{X}}_{3} \hat{\mathbf{A}}_{3}^{k}\right\|_{\text{F}}^{2}+\frac{\rho}{2}\left\|\hat{\mathbf{X}}_{3}-\hat{\mathbf{O}}_{3}^{k}\right\|_{\text{F}}^{2}\\
&+\left\langle\Lambda^{p}, \mathbf{B} \hat{\mathbf{X}}_{3}+\mathbf{C} \mathbf{U}^{p}\right\rangle+\frac{\beta}{2}\left\|\mathbf{B} \hat{\mathbf{X}}_{3}+\mathbf{C} \mathbf{U}^{p}\right\|_{\text{F}}^{2}.
\end{aligned}
\end{equation}
Then, the solution of (\ref{equation_X3_subsub}) can be obtained by using the classical Sylvester matrix equation
\begin{equation}
\label{equation_sy}
\hat{\mathbf{X}}_{3}(\hat{\mathbf{A}}_{3}^{k} (\hat{\mathbf{A}}_{3}^{k})^{T})+\beta \mathbf{B}^{T} \mathbf{B} \hat{\mathbf{X}}_{3}+\rho \hat{\mathbf{X}}_{3}=\rho \hat{\mathbf{O}}_{3}^{k}+\hat{\mathbf{Y}}_{(3)}^{k}(\hat{\mathbf{A}}_{3}^{k})^{T}-\mathbf{B}^{T} \Lambda^{p}-\beta \mathbf{B}^{T} \mathbf{C} \mathbf{U}^{p}.
\end{equation}
By using the Kronecker product notations, (\ref{equation_sy}) can be rewritten as:
\begin{equation}
\label{equation_X3_KP}
\begin{aligned}
&\left(\hat{\mathbf{A}}_{3}^{k}\left(\hat{\mathbf{A}}_{3}^{k}\right)^{T} \otimes \mathbf{I}+\beta \mathbf{I} \otimes \mathbf{B}^{T} \mathbf{B}+\rho_{1} \mathbf{I} \otimes \mathbf{I}) \operatorname{vec}\left(\hat{\mathbf{X}}_{3}\right)\right.\\
&=\operatorname{vec}\left(\rho \hat{\mathbf{O}}_{3}^{k}+\hat{\mathbf{Y}}_{(3)}^{k}(\hat{\mathbf{A}}_{(3)}^{k})^{T}-\mathbf{B}^{T} \Lambda^{p}-\beta \mathbf{B}^{T} \mathbf{C} \mathbf{U}^{p}\right)
\end{aligned}
\end{equation}
where $\operatorname{vec}(.)$ refers to a vector by lexicographical ordering of the entries in a matrix.
Using SVD of $\hat{\mathbf{A}}_{3}^{k}$, i.e., $\hat{\mathbf{A}}_{3}^{k}=\mathbf{P} \Sigma \mathbf{Q}^{*}$,
and the Fourier decomposition of $ \mathbf{B}^{T} \mathbf{B}$ with periodic boundary condition, i.e., $\mathbf{B}^{T} \mathbf{B}=\mathbf{F}^{*} \Psi^{2} \mathbf{F}$,
we can solve the problem ( \ref{equation_X3_KP}) efficiently.
Then, (\ref{equation_X3_KP}) can be rewritten as:
\begin{equation}
\begin{aligned}
&\left(\mathbf{P} \otimes \mathbf{F}^{*}\right)\left(\Sigma^{2} \otimes \mathbf{I}+\beta \mathbf{I} \otimes \Psi^{2}+\rho \mathbf{I} \otimes \mathbf{I}\left(\mathbf{P}^{*} \otimes \mathbf{F}\right) \operatorname{vec}\left(\hat{\mathbf{X}}_{3}\right)\right.\\
&=\operatorname{vec}\left(\rho \hat{\mathbf{O}}_{3}^{k}+\hat{\mathbf{Y}}_{(3)}^{k}\left(\hat{\mathbf{A}}_{3}^{k}\right)^{T}-\mathbf{B}^{T} \Lambda^{p}-\beta \mathbf{B}^{T} \mathbf{C} \mathbf{U}^{p}\right).
\end{aligned}
\end{equation}
The solution $\operatorname{vec}\left(\hat{\mathbf{X}}_{3}\right)$ is explicitly expressed as:
\begin{equation}
\label{equation:solution_X3}
\begin{aligned}
\operatorname{vec}\left(\hat{\mathbf{X}}_{3}\right)=&\left(\mathbf{P} \otimes \mathbf{F}^{*}\right) {\left(\Sigma^{2} \otimes \mathbf{I}+\beta \mathbf{I} \otimes \Psi^{2}+\rho \mathbf{I} \otimes \mathbf{I}\right)}^{-1}\left(\mathbf{P}^{*} \otimes \mathbf{F}\right)\\
&\cdot \operatorname{vec}\left(\rho \hat{\mathbf{O}}_{3}^{k}+\hat{\mathbf{Y}}_{(3)}^{k}\left(\hat{\mathbf{A}}_{3}^{k}\right)^{T}-\mathbf{B}^{T} \Lambda^{p}-\beta \mathbf{B}^{T} \mathbf{C} \mathbf{U}^{p}\right).
\end{aligned}
\end{equation}

With other variables fixed, the minimization subproblem for $\mathbf{U}$ can be deduced from (\ref{equation_alm_X3}) as follows:
\begin{equation}
\label{equation_W}
\mathbf{U}^{p+1}=\arg\min_{\mathbf{U}} \mu \sum_{i=1}^{s_{3}} \sum_{j=1}^{r_{3}}\left\|\mathbf{U}_{i, j}\right\|_{2}+\frac{\beta}{2}\left\|\mathbf{B}\hat{\mathbf{X}}_3^{k+1, p+1}+\mathbf{C} \mathbf{U}+\frac{\Lambda^{p}}{\beta}\right\|_{\text{F}}^{2}.
\end{equation}
Its solution can be transformed into solving $s_{3}r_{3}$ two-variable minimization problems independently as follows:
\begin{equation}
\label{equation_W_2v}
\begin{aligned}
\arg\min_{\mathbf{U}_{1},\mathbf{U}_{2}} & \mu \sqrt{\left|\left(\mathbf{U}_{1}\right)_{i, j}\right|^{2}+\left|\left(\mathbf{U}_{2}\right)_{i, j}\right|^{2}}+\frac{\beta}{2}\left[\left(\mathbf{U}_{1}\right)_{i, j}-\left(\mathbf{D}_{1} \hat{\mathbf{X}}_{3}^{k+1, p+1}\right)_{i, j}-\frac{1}{\beta}\left(\Lambda_{1}^{p}\right)_{i, j}\right]^{2}\\
&\quad+\frac{\beta}{2}\left[\left(\mathbf{U}_{2}\right)_{i, j}-\left(\mathbf{D}_{2} \hat{\mathbf{X}}_{3}^{k+1, p+1}\right)_{i, j}-\frac{1}{\beta}\left(\Lambda_{2}^{p}\right)_{i, j}\right]^{2}.
\end{aligned}
\end{equation}
The solution of (\ref{equation_W_2v}) can be obtained by using the well-known 2-D shrinkage formula
\begin{equation}
\label{equation_solution_U}
\left[\left(\mathbf{U}_{1}\right)_{i, j},\left(\mathbf{U}_{2}\right)_{i, j}\right]=\max \left\{\left\|\mathbf{T}_{i, j}\right\|_{2}-\frac{\mu}{\beta}, 0\right\} \frac{\mathbf{T}_{i, j}}{\left\|\mathbf{T}_{i, j}\right\|_{2}},
\end{equation}
where $\mathbf{T}_{i, j}=\left[\left(\mathbf{D}_{1} \hat{\mathbf{X}}_{3}^{k+1, p+1}\right)_{i, j}+\frac{1}{\beta}\left(\Lambda_{1}^{p}\right)_{i, j},\left(\mathbf{D}_{2} \hat{\mathbf{X}}_{3}^{k+1, p+1}\right)_{i, j}+\frac{1}{\beta}\left(\Lambda_{2}^{p}\right)_{i, j}\right]$, $1\leq i \leq s_{3}, 1 \leq j \leq r_{3}$;
we assign $0 \cdot(0 / 0)=0$, as stated in \cite{MFTV}.

After solving the two sub-problems with respect to $\hat{\mathbf{X}}_3$ and $\mathbf{U}$,
the Lagrangian multipliers $\Lambda=\left(\Lambda_{1} , \Lambda_{2}\right)^{T}$ can be updated in parallel as
\begin{equation}
\label{equation_solution_LL}
\Lambda^{p+1}=\Lambda^{p}+\beta\left(\mathbf{B} \hat{\mathbf{X}}_{3}^{k+1, p+1}+C \mathbf{U}^{p+1}\right).
\end{equation}

\subsubsection{Update $\mathbf{A}_n$ with fixing others}

The $\mathbf{A}_n$-sub-problem in (\ref{equation:XAY}) can be written as follows:
\begin{equation}
\label{equation_A_aux}
\begin{aligned}
\mathbf{A}_n^{k+1}&=\arg\min_{\mathbf{A}_n} \sum_{n=1}^{N} (\frac{\alpha_{n}}{2}\left\|\mathbf{Y}_{(n)}-\mathbf{A}_{n} \mathbf{X}_{n}\right\|_{\text{F}}^{2}+\lambda_n \left\|\mathbf{A}_{n}\right\|_{\text{*}}+\frac{\rho_n}{2}\left\|\mathbf{A}_n-\mathbf{A}_n^{k}\right\|_{\text{F}}^{2}).
\end{aligned}
\end{equation}
By introducing an auxiliary variable, (\ref{equation_A_aux}) can be rewritten as
\begin{equation}
\label{equation_A_alm}
\begin{aligned}
&\arg\min_{\mathbf{A}_n} \sum_{n=1}^{N} (\frac{\alpha_{n}}{2}\left\|\mathbf{Y}_{(n)}-\mathbf{A}_{n} \mathbf{X}_{n}\right\|_{\text{F}}^{2}+\lambda_n \left\|\mathbf{J}_{n}\right\|_{\text{*}}
+\frac{\rho_n}{2}\left\|\mathbf{A}_n-\mathbf{A}_n^{k}\right\|_{\text{F}}^{2})\\
&s.t., \mathbf{A}_{n}=\mathbf{J}_{n}.
\end{aligned}
\end{equation}
By the ALM method, the problem (\ref{equation_A_alm}) can also be reformulated as
\begin{equation}
\label{equation:A_alm}
\begin{aligned}
\arg\min_{\mathbf{A}_n, \mathbf{J}_n} & \sum_{n=1}^{N} (\frac{\alpha_{n}}{2}\left\|\mathbf{Y}_{(n)}-\mathbf{A}_{n} \mathbf{X}_{n}\right\|_{\text{F}}^{2}+\lambda_n \left\|\mathbf{J}_{n}\right\|_{\text{*}}
+\frac{\rho_n}{2}\left\|\mathbf{A}_n-\mathbf{A}_n^{k}\right\|_{\text{F}}^{2}\\
&+\left\langle\Gamma_{n}^\mathbf{A}, \mathbf{A}_n-\mathbf{J}_n\right\rangle+\frac{\rho_{n}}{2}\left\|\mathbf{A}_{n}-\mathbf{J}_{n}\right\|_{\text{F}}^{2}),
\end{aligned}
\end{equation}
where $\Gamma_{n}^\mathbf{A}$ is the Lagrangian multiplier.

Firstly, with other variables fixed, the minimization subproblem for $\mathbf{J}_n$ can be deduced from (\ref{equation:A_alm}) as follows:
\begin{equation}
\displaystyle \mathbf{J}_n^{k+1}= \arg \min_{\mathbf{J}_n} \lambda_n \left\|\mathbf{J}_{n}\right\|_{\text{*}}+\frac{\rho_{n}}{2}\left\|\mathbf{A}_{n}^{k}-\mathbf{J}_{n}+\Gamma_{n}^{\mathbf{A}}/\rho_n\right\|_{\text{F}}^{2}.
\end{equation}
Its solution can also be obtained by SVT operator (\ref{SVT})
\begin{equation}
\label{equation_solution_Jn}
\begin{aligned}
\mathbf{J}_n^{k+1}=\operatorname{SH}_{\frac{\lambda_n}{\rho_{n}}}(\mathbf{A}_{n}^{k}+\Gamma_{n}^{\mathbf{A}}/\rho_n ), n=1,2, \cdots, N.
\end{aligned}
\end{equation}

Secondly, with other variables fixed, the minimization subproblem for $\mathbf{A}_n$ can be deduced from (\ref{equation:A_alm}) as follows:
\begin{equation}
\label{equationforA}
\begin{aligned}
\mathbf{A}_n^{k+1}= \arg\min_{\mathbf{A}_n} \sum_{n=1}^{N} ( \frac{\alpha_{n}}{2}\left\|\mathbf{Y}_{(n)}-\mathbf{A}_{n} \mathbf{X}_{n}\right\|_{\text{F}}^{2}+\rho_{n}\left\|\mathbf{A}_{n}-\frac{\mathbf{J}_{n}^{k+1}-\Gamma_{n}^\mathbf{A}/\rho_n+\mathbf{A}_n^{k}}{2}\right\|_{\text{F}}^{2}).
\end{aligned}
\end{equation}
It is also convex and has the following closed-form solution
\begin{equation}
\label{equation_solution_An}
\begin{array}{r}
\mathbf{A}_{n}^{k+1}=\left(\mathbf{X}_{(n)}^{k}\left(\mathbf{X}_{n}^{k+1}\right)^{T}+2\rho_n (\frac{\mathbf{J}_{n}^{k+1}-\Gamma_{n}^\mathbf{A}/\rho_n+\mathbf{A}_n^{k}}{2})\right)\left(\mathbf{X}_{n}^{k+1}\left(\mathbf{X}_{n}^{k+1}\right)^{T}+2\rho_{n} \mathbf{I}_{n}\right)^{\dagger}, \\
n=1,2,\cdots, N.
\end{array}
\end{equation}

Finally, the Lagrangian multiplier can be updated by the following equations
\begin{equation}
\label{equation:Lambda_1}
\Gamma_{n}^{\mathbf{A}} = \Gamma_{n}^{\mathbf{A}} + \mathbf{A}_n-\mathbf{J}_n.
\end{equation}

\subsubsection{Update $\mathcal{Y}$ with fixing others}

With other variables fixed, the minimization subproblem for $\mathbf{Y}_{(n)}$ in (\ref{equation:XAY}) can be written as
\begin{equation}
\begin{aligned}
\mathbf{Y}_{(n)}^{k+1} = \arg\min_{\mathbf{Y}_{(n)}} & \sum_{n=1}^{N} ( \frac{\alpha_{n}}{2}\left\|\mathbf{Y}_{(n)}-\mathbf{A}_{n} \mathbf{X}_{n}\right\|_{\text{F}}^{2}+\frac{\rho}{2}\left\|\mathcal{Y}-\mathcal{Y}^{k}\right\|_{\text{F}}^{2} \\
& s.t., \mathcal{P}_{\Omega}(\mathcal{Y})=\mathcal{F}.
\end{aligned}
\end{equation}
Then, the update of $\mathcal{Y}_{k+1}$ can be written explicitly as
\begin{equation}
\label{equation_solution_Y}
\begin{array}{l}
\displaystyle \mathcal{Y}^{k+1}=P_{{\Omega}^c}\left(\sum_{n=1}^{N} \alpha_{n} \text { fold }_{n}\left(\frac{\mathbf{A}_{n}^{k+1} \mathbf{X}_{n}^{k+1}+\rho_n \mathbf{Y}_{(n)}^{k}}{1+\rho_n}\right)\right)+\mathcal{F},
\end{array}
\end{equation}
where $\mathcal{F}$ is the observed data;
$P_{{\Omega}}$ is an operator defined in subsection \ref{operators}.

The above proposed algorithm is applicable to the proposed model-1 and model-2,
due to all variables of model-1 and model-2 are updated in the same way except $\mathbf{X}$.
Specifically, model-2 has one more regularizer applied to $\mathbf{X}_3$ than model-1.
Therefore, model-1 updates $\mathbf{X}_n, n=1,2, \cdots, N$ according to (\ref{equation:solution_Xn});
model-2 updates $\mathbf{X}_n(n\not=3)$ according to (\ref{equation:solution_Xn}),
while updates $\mathbf{X}_3$ according to (\ref{equation:solution_X3}), (\ref{equation_solution_U}) and (\ref{equation_solution_LL}).

\subsection{ Complexity and Converge Analysis}

In this subsection, the proposed algorithm for the proposed model-1 and model-2 are summarized as Algorithm \ref{algorithm:A1} and \ref{algorithm:A2}.
Further, we discuss the complexity and convergence of the proposed algorithms.

\subsubsection{Complexity Analysis}

The cost of computing $\mathbf{X}_{n}$ is $O\left(I_{n} r_{n}^{2}+I_{n} r_{n} s_{n}+r_{n}^{2} s_{n}\right)$;
calculating $\mathbf{Z_n}$ has a complexity of $O\left( \Pi_{j \neq n} I_{j} \times r_{n}^2 \right)$;
the complexity of updating $\mathbf{J}_n$ is $O\left(I_{n} r_{n}^2\right)$;
calculating $\mathbf{A}_{n}$ has a complexity of $O\left(I_{n} r_{n}^{2}+I_{n} r_{n} s_{n}+r_{n}^{2} s_{n}\right)$
and calculating $[\mathbf{U}_1, \mathbf{U}_2]$ has a complexity of $O\left(s_{3} r_{3}\right)$;
the complexity of updating $\operatorname{vec}\left(\hat{\mathbf{X}}_{3}\right)$ is  $O\left(2 s_{3} r_{3}^{2}+s_{3} r_{3} \log s_{3}\right)$;
calculating $\mathcal{Y}$ has a complexity of $O\left(r_{1} I_{1} s_{1}+\cdots+r_{N} I_{N} s_{N}\right)$.
Then, the total complexity of the proposed algorithms can be obtained by counting the complexity of the above variables.
For easily viewing, we list the total complexity of the proposed model-1 and model-2 in (\ref{equation:complexity_model1}) and (\ref{equation:complexity_model2}), respectively.
\begin{equation}
\label{equation:complexity_model1}
O(\sum_{n \neq 3}(3I_{n} r_{n}^2+\Pi_{j \neq n} I_{j} \times r_{n}^2+3 I_{n} r_{n} S_{n}+2 r_{n}^{2} s_{n}))
\end{equation}
\begin{equation}
\label{equation:complexity_model2}
O(I_{3} r_{3}^{2}+2 l_{3} r_{3} s_{3}+3 r_{3}^{2} s_{3}+r_{3} s_{3} \log s_{3}+\sum_{n \neq 3}(3I_{n} r_{n}^2+\Pi_{j \neq n} I_{j} \times r_{n}^2+3 I_{n} r_{n} S_{n}+2 r_{n}^{2} s_{n}))
\end{equation}

\begin{algorithm}[!t]
	\caption{:Algorithm for the proposed model-1.} \label{algorithm:A1}
	\begin{algorithmic}[1]
		\Require
		The observed tensor $\mathcal{F}$;
		The set of index of observed entries $\Omega$;
		The given $n$-rank, $r = (r_1, r_2, r_3)$;
		stopping criterion $\varepsilon.$
		\Ensure		
		The completed tensor.		
		\State Initialize:  $\mathbf{X}_n^0=\mathbf{Z}_n^0=\mathbf{0}, \mathbf{A}_n^0=\mathbf{J}_n^0=\mathbf{0}, \Gamma_{n}^\mathbf{X}=\mathbf{0},
		\Gamma_{n}^\mathbf{A}=\mathbf{0},  n=1,2,\cdots, N$; $ \mu_{\max }=10^{6}, \rho=1.5,$ $\mathcal{Y}=\mathcal{P}_{\Omega}(\mathcal{F})$, and $k=0$.
		\State Repeat until convergence:		
		\State Update $\mathbf{X}, \mathbf{Z}, \mathbf{A}, \mathbf{J}, \mathcal{Y},  \Gamma^{\mathbf{X}}, \Gamma^{\mathbf{A}}$ via
		
		1st step: Update $\mathbf{Z}_n$ via	(\ref{equation_solution_Zn})
		
		2nd step: Update $\mathbf{X}_n, n=1,2, \cdots, N,$ via	(\ref{equation:solution_Xn})
		
		3rd step: Update $\mathbf{A}_n$ via (\ref{equation_solution_An})
		
		4th step: Update $\mathbf{J}_n$ via	(\ref{equation_solution_Jn})
		
		5th step: Update $\mathcal{Y}$ via (\ref{equation_solution_Y})
		
		6th step: Update the parameter via 	(\ref{equation:Lambda_0}), (\ref{equation:Lambda_1})
		
		\State Check the convergence condition.
	\end{algorithmic}
\end{algorithm}

\subsubsection{Convergence Analysis}

In this subsection, the convergence of the proposed algorithms is proved theoretically by using the block successive upper-bound minimization (BSUM) \cite{BSUM}.
The BSUM is an alternative inexact block coordinate descent method which is proposed recently.
It is designed for non-smooth optimization problem.

\noindent \textbf{Lemma 1} \cite{BSUM, MFTV}. Given the problem $\arg\min f(x)$, s.t. $x \in \mathcal{X},$ where $\mathcal{X}$ is the feasible set.
Assume $h\left(x, x^{k-1}\right)$ is an approximation of $f(x)$ at the $(k-1)$th iteration, which satisfied the following conditions:
\begin{equation}
\begin{array}{l}
1) \quad h_{i}\left(y_{i}, y\right)=f(y), \forall y \in \mathcal{X}, \forall i; \\
2) \quad h_{i}\left(x_{i}, y\right) \geq f\left(y_{1}, \ldots, y_{i-1}, x_{i}, y_{i+1}, \ldots, y_{n}\right), \forall x_{i} \in \mathcal{X}_{i}, \forall y \in \mathcal{X}, \forall i_{i};\\
3) \quad \left.h_{i}^{\prime}\left(x_{i}, y ; d_{i}\right)\right|_{x_i=y_i}=f^{\prime}(y ; d), v_{i}=\left(0, \ldots, d_{i} \ldots 0\right) \text { s.t. } y_{i}+d_{i} \in \mathcal{X}_{i}, \forall i;\\
4) \quad 	h_{i}\left(x_{i}, y\right) \text{is continuous in} \left(x_{i}, y\right), \forall i;
\end{array}
\end{equation}
where $h_{i}\left(x_{i}, y\right)$ is the sub-problem with respect to the $i$th block and $f^{\prime}(y ; d)$ is the direction derivative of $f$ at the point $y$ in direction $d$.
Suppose $h_{i}\left(x_{i}, y\right)$ is quasi-convex in $x_{i}$ for $i=1,2, \cdots,  n$.
Furthermore, assume that each sub-problem $\operatorname{argmin} h_i\left(x_{i}, x^{k-1}\right),$ s.t. $x \in \mathcal{X}_{i}$ has a unique solution for any point $x^{k-1} \in \mathcal{X} .$
Then, the iterates generated by the BSUM algorithm converge to the set of coordinatewise minimum of $f$.

\begin{algorithm}[!t]
	\caption{:Algorithm for the proposed model-2.} \label{algorithm:A2}
	\begin{algorithmic}[1]
		\Require
		The observed tensor $\mathcal{F}$; The set of index of observed entries $\Omega$;
		The given $n$-rank, $r = (r_1, r_2, r_3)$; stopping criterion $\varepsilon.$
		\Ensure		
		Output: The completed tensor;		
		\State \textbf{Initialize}:  $\mathbf{A}_{n}^{0}=\operatorname{rand}\left(I_{n} \times r_{n}\right), \mathbf{X}_{n}^{0}=\operatorname{rand}\left(r_{n} \times \prod_{m=1, m \neq n}^{N} I_{m}\right),(n=1,2,\ldots, N), \mathcal{Y}=\mathcal{P}_{\Omega}(\mathcal{F})$.
		
		\Repeat
		
		1st step: Update $\mathbf{Z}_n$ via	(\ref{equation_solution_Zn})
		
		2nd step: Update $\mathbf{X}_n, n=1,2,4,5, \cdots, N,$ via	(\ref{equation:solution_Xn})	
		
		3rd step: Update $\mathbf{X}_3$ via
		\Repeat
		
		3-1st step: Update $\mathbf{X}_3$ via (\ref{equation:solution_X3})
		
		3-2nd step: Update $\mathbf{U}$ via (\ref{equation_solution_U})
		
		3-3rd step: Update $\Lambda$ via (\ref{equation_solution_LL})
		
		\Until{converged}
		
		4th step: Update $\mathbf{J}_n$ via	(\ref{equation_solution_Jn})
		
		5th step: Update $\mathbf{A}_n$ via (\ref{equation_solution_An})
		
		6th step: Update $\mathcal{Y}$ via (\ref{equation_solution_Y})
		
		7th step: Update the parameter via 	(\ref{equation:Lambda_0}), (\ref{equation:Lambda_1})
		\Until{converged }
	\end{algorithmic}
\end{algorithm}


\noindent \textbf{Theorem 1.} The iterates generated by (\ref{equation_original_PPA}) converge to the set of coordinatewise minimizers.

\noindent \textbf{Proof.} According to the notations in (\ref{equation_original_PPA}) and (\ref{equation:XAY}), we give the notions for convenience
\begin{equation}
\left\{\begin{array}{l}
g(\mathcal{S}, \mathcal{S}^k)=  f\left(\mathcal{S}\right)+\frac{\rho}{2}\left\|\mathcal{S}-\mathcal{S}^k\right\|_{\text{F}}^{2}, \\
g_1\left(\mathbf{X}, \mathcal{S}_{1}^{k}\right) = f\left(\mathbf{X}, \mathbf{A}^{k}, \mathcal{Y}^{k}\right)+\frac{\rho}{2}\left\|\mathbf{X}-\mathbf{X}^{k}\right\|_{\text{F}}^{2}, \\
g_2\left(\mathbf{A}, \mathcal{S}_{2}^{k}\right) = f\left(\mathbf{X}^{k+1}, \mathbf{A}, \mathcal{Y}^{k}\right)+\frac{\rho}{2}\left\|\mathbf{A}-\mathbf{A}^{k}\right\|_{\text{F}}^{2}, \\
g_3\left(\mathcal{Y}, \mathcal{S}_{3}^{k}\right) = f\left(\mathbf{X}^{k+1}, \mathbf{A}^{k+1}, \mathcal{Y}\right)+\frac{\rho}{2}\left\|\mathcal{Y}-\mathcal{Y}^{k}\right\|_{\text{F}}^{2}.
\end{array}\right.	
\end{equation}
It is easy to verify that $g\left(\mathcal{S}, \mathcal{S}^{k}\right)$ is an approximation and a global upper bound of $f(\mathcal{S})$ at the $k$th iteration, which satisfies the following conditions:
\begin{equation}
\begin{array}{l}
1) \quad 	g_{i}\left(\mathcal{S}_{i}, \mathcal{S}\right)=f(\mathcal{S}), \forall \mathcal{S}, i=1,2,3; \\
2) \quad 	g_{i}\left(\bar{\mathcal{S}}_{i}, \mathcal{S}\right) \geq f\left(\mathcal{S}_{1}, \ldots, \bar{\mathcal{S}_{i}}, \ldots, \mathcal{S}_{3}\right), \forall \bar{\mathcal{S}}_{i}, \forall \mathcal{S}, i=1,2,3; \\
3) \quad 	\left.g_{i}^{\prime}\left(\bar{\mathcal{S}}_{i}, \mathcal{S} ; \mathbf{M}_{i}\right)\right|_{\bar{\mathcal{S}}_{i}=\mathcal{S}_{i}}=f^{\prime}\left(\mathcal{S} ; \mathbf{M}^i\right), \forall \mathbf{M}^{i}=\left(0, \ldots, \mathbf{M}_{i},\ldots, 0\right); \\
4) \quad 	g_{i}\left(\bar{\mathcal{S}}_{i}, \mathcal{S}\right) \text{is continuous in} \left(\bar{\mathcal{S}}_{i}, \mathcal{S} \right), i=1,2,3;
\end{array}	
\end{equation}
where $\mathcal{S}=(\mathcal{S}_{1},\mathcal{S}_{2},\mathcal{S}_{3})=(\mathbf{X}, \mathbf{A}, \mathcal{Y})$.
In addition, the sub-problem $g_{i}(i=1,2,3)$ is strictly convex
with respect to $\mathbf{X}, \mathbf{A}$ and $\mathcal{Y}$ respectively and thus each sub-problem has a unique solution.
Therefore, all assumptions in \textbf{Lemma 1} are satisfied.
According to the conclusion of \textbf{Lemma 1}, the \textbf{Theorem 1} is valid, and the proposed algorithms are theoretically convergent.

\section{Numerical experiments}

In order to verify the effectiveness of the proposed model-1 and model-2,
we carry out lots of experiments on three types of public tensor data sets, i.e., video data, MRI data and hyperspectral image data,
which have been frequently used to interpret the tensor completion performance of different models.
Four different completion models are selected as comparison methods, i.e.,
TMac \cite{Tmac}, TV based MF-TV method \cite{MFTV}, single nuclear norm based TNN method \cite{TNN} and partial sum of tubal nuclear norm based PSTNN method \cite{PSTNN}.

To accurately evaluate the performance of the models, we mainly use two types of standards for evaluation.
The first is the visual evaluation of the restored data, which is a qualitative evaluation standard.
The second is the five quantitative picture quality indices (PQIs),
including the peak signal-to-noise ratio (PSNR) \cite{PSNR},
structural similarity index (SSIM) \cite{SSIM},
feature similarity (FSIM) \cite{FSIM},
erreur relative globale adimensionnelle de synth\`ese (ERGAS) \cite{EGRAS},
the mean the spectral angle mapper (SAM) \cite{SAM}.
Larger PSNR, SSIM, FSIM and smaller ERGAS, SAM are, the better the restoration performance of the corresponding model is.
Since the experimental datasets are all third-order tensors,
the PQIs for each frontal slice in the restored tensor are first calculated,
and then the mean of these PQIs are finally used to evaluate the performance of the models.
All experiments were performed on MATLAB 2018b, the CPU of the computer is Inter core i7@2.2GHz and the memory is 64GB.

For a tensor $\mathcal{Y} \in \mathbb{R}^{I_1 \times \ldots \times I_N}$,
let $S_{\text{number}}$ denote the number of sampled entries in its index set $\Omega$.
Then the sampling ratio (SR) can be defined as:
\begin{equation}
\mathrm{SR}=\frac{S_{\text{number}}}{\prod_{n=1}^{N} I_{n}},
\end{equation}where the sampled entries are chosen randomly from a tensor $\mathcal{Y}$ by a uniform distribution.
In the proposed algorithms for model-1 and model-2, the inputs include the observed tensor $\mathcal{F} \in \mathbb{R}^{I_1 \times I_2 \times I_3}$,
the stopping criteria $\epsilon$, the regularized parameters $\alpha, \beta, \lambda, \tau,$ and the penalty parameter $\beta$.
All parameters are empirically.
Specifically, the stopping criterion $\epsilon$ and the weights $\alpha_{i}(i=1,2,3)$ of the proposed model-1 and model-2 are set to be $10^{-5}$ and $1/3$ for all experiments;
the regularization parameter $\mu$ and the penalty parameter $\beta$ for model-2 are set as 0.5 and 10, respectively;
finally, the proximal parameter $\rho$ and regularized parameters $\lambda, \tau$ are all set as 0.1 for all experiments of model-1 and model-2.

\subsection{Video}

In this part, the proposed model is applied to two video datasets to verify the performance of the model.
The two video datasets are video dataset "suzie" and “hall”\footnote{http://trace.eas.asu.edu/yuv/}, both of which are colored using YUV format.
Their sizes are 144 $\times$ 176 $\times$ 150.
The sampling rates are set as 5\%, 10\% and 20\%.

\begin{figure*}[!t]	
\centering			
\subfloat[Original]{\includegraphics[width=0.23\linewidth]{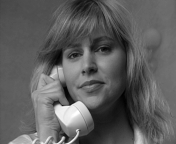}}%
\hfil	
\subfloat[95\% Masked]{\includegraphics[width=0.23\linewidth]{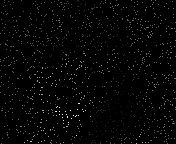}}%
\hfil					
\subfloat[our model-2]{\includegraphics[width=0.23\linewidth]{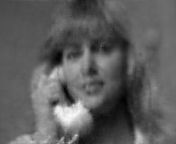}}%
\hfil
\subfloat[our model-1]{\includegraphics[width=0.23\linewidth]{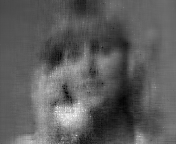}}%
\hfil
\subfloat[MF-TV]{\includegraphics[width=0.23\linewidth]{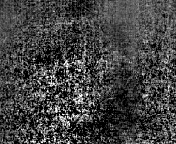}}%
\hfil					
\subfloat[Tmac]{\includegraphics[width=0.23\linewidth]{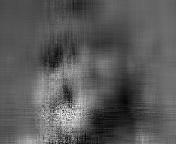}}%
\hfil
\subfloat[PSTNN]{\includegraphics[width=0.23\linewidth]{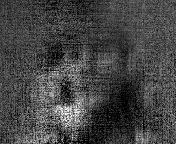}}%
\hfil
\subfloat[TNN]{\includegraphics[width=0.23\linewidth]{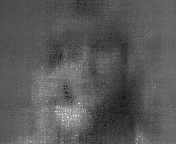}}%
\caption{One slice of the recovered video for “suzie” by our model-1 and model-2, MF-TV, Tmac, PSTNN and TNN.  The sampling rate is 5\%.}
\label{figure_video_sr0.05}
\end{figure*}

\begin{figure*}[!t]	
\centering			
\subfloat[Original]{\includegraphics[width=0.23\linewidth]{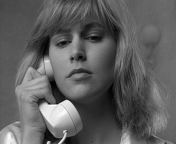}}%
\hfil	
\subfloat[90\% Masked]{\includegraphics[width=0.23\linewidth]{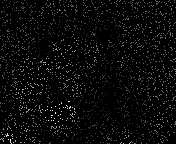}}%
\hfil					
\subfloat[our model-2]{\includegraphics[width=0.23\linewidth]{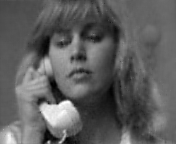}}%
\hfil
\subfloat[our model-1]{\includegraphics[width=0.23\linewidth]{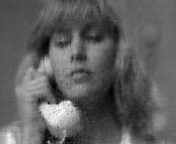}}%
\hfil
\subfloat[MF-TV]{\includegraphics[width=0.23\linewidth]{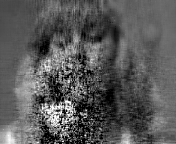}}%
\hfil					
\subfloat[Tmac]{\includegraphics[width=0.23\linewidth]{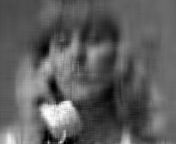}}%
\hfil
\subfloat[PSTNN]{\includegraphics[width=0.23\linewidth]{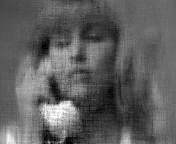}}%
\hfil
\subfloat[TNN]{\includegraphics[width=0.23\linewidth]{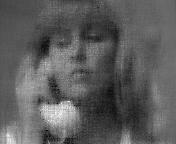}}%
\caption{One slice of the recovered video for “suzie” by our model-1, model-2, MF-TV, Tmac, PSTNN and TNN. The sampling rate is 10\%.}
\label{figure_video_sr0.1}
\end{figure*}

\begin{figure*}[!t]	
\centering			
\subfloat[Original]{\includegraphics[width=0.23\linewidth]{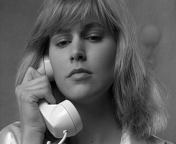}}%
\hfil	
\subfloat[80\% Masked]{\includegraphics[width=0.23\linewidth]{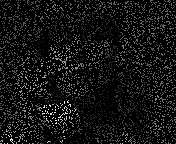}}%
\hfil					
\subfloat[our model-2]{\includegraphics[width=0.23\linewidth]{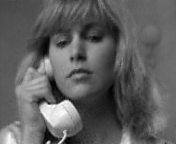}}%
\hfil
\subfloat[our model-1]{\includegraphics[width=0.23\linewidth]{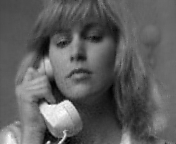}}%
\hfil
\subfloat[MF-TV]{\includegraphics[width=0.23\linewidth]{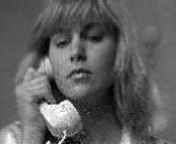}}%
\hfil					
\subfloat[Tmac]{\includegraphics[width=0.23\linewidth]{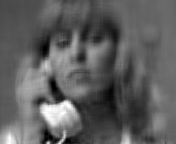}}%
\hfil
\subfloat[PSTNN]{\includegraphics[width=0.23\linewidth]{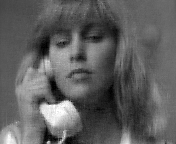}}%
\hfil
\subfloat[TNN]{\includegraphics[width=0.23\linewidth]{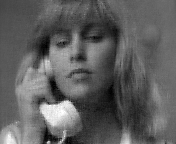}}%
\caption{One slice of the recovered video for “suzie” by our model-1 and model-2, MF-TV, Tmac, PSTNN and TNN.  The sampling rate is 20\%.}
\label{figure_video_sr0.2}
\end{figure*}

\begin{figure*}[!t]	
\centering			
\subfloat[Original]{\includegraphics[width=0.23\linewidth]{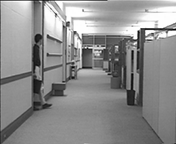}}%
\hfil	
\subfloat[95\% Masked]{\includegraphics[width=0.23\linewidth]{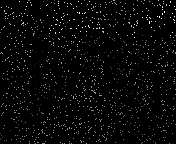}}%
\hfil					
\subfloat[our model-2]{\includegraphics[width=0.23\linewidth]{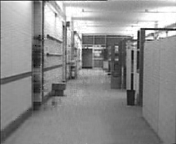}}%
\hfil
\subfloat[our model-1]{\includegraphics[width=0.23\linewidth]{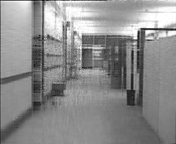}}%
\hfil
\subfloat[MF-TV]{\includegraphics[width=0.23\linewidth]{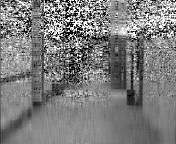}}%
\hfil					
\subfloat[Tmac]{\includegraphics[width=0.23\linewidth]{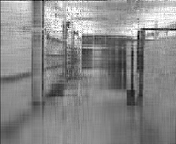}}%
\hfil
\subfloat[PSTNN]{\includegraphics[width=0.23\linewidth]{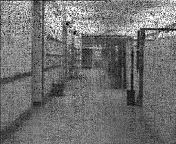}}%
\hfil
\subfloat[TNN]{\includegraphics[width=0.23\linewidth]{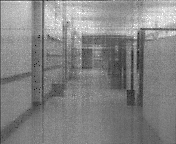}}%
\caption{One slice of the recovered video for “hall” by our model-1 and model-2, MF-TV, Tmac, PSTNN and TNN.  The sampling rate is 5\%.}
\label{figure_hall_sr0.05}
\end{figure*}

\begin{figure*}[!t]	
\centering			
\subfloat[Original]{\includegraphics[width=0.23\linewidth]{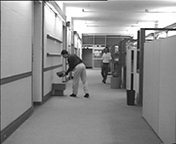}}%
\hfil	
\subfloat[90\% Masked]{\includegraphics[width=0.23\linewidth]{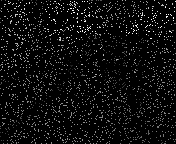}}%
\hfil					
\subfloat[our model-2]{\includegraphics[width=0.23\linewidth]{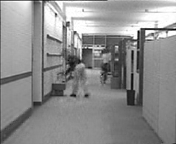}}%
\hfil
\subfloat[our model-1]{\includegraphics[width=0.23\linewidth]{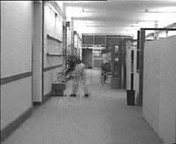}}%
\hfil
\subfloat[MF-TV]{\includegraphics[width=0.23\linewidth]{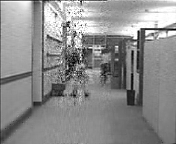}}%
\hfil					
\subfloat[Tmac]{\includegraphics[width=0.23\linewidth]{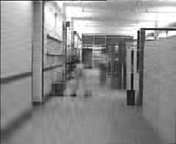}}%
\hfil
\subfloat[PSTNN]{\includegraphics[width=0.23\linewidth]{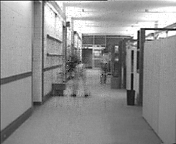}}%
\hfil
\subfloat[TNN]{\includegraphics[width=0.23\linewidth]{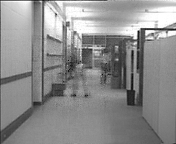}}%
\caption{One slice of the recovered video “hall” by our model-1 and model-2, MF-TV, Tmac, PSTNN and TNN.  The sampling rate is 10\%.}
\label{figure_hall_sr0.1}
\end{figure*}

For quantitative comparison, Table \ref{table_video_suzie} and Table \ref{table_video_hall} list the PQIs of all the compared models in the three sampling rates.
The best results for each PQI are marked in bold.
It is clear from Table \ref{table_video_suzie} and Table \ref{table_video_hall} that in all SR cases our model-2 obtain the best results,
and our model-1 obtain the suboptimal results compared to other compared methods.
For visual evaluation, we show one frontal slice of the recovered results with different random sampling rates
in Fig. \ref{figure_video_sr0.05},
Fig. \ref{figure_video_sr0.1},
Fig. \ref{figure_video_sr0.2},
Fig. \ref{figure_hall_sr0.05} and Fig. \ref{figure_hall_sr0.1}.
Compared to other models, it can be seen that the results of our models are closest to the original reference images, especially at low sampling rates.
Specifically, as shown in Fig. \ref{figure_hall_sr0.1}, Fig. \ref{figure_video_sr0.05} and Fig. \ref{figure_video_sr0.1},
when the sampling rate is 0.05 and 0.1, the advantages of the proposed models are most obvious.
The proposed models restore most of the structural information of the image,
while the image restored by the competitive method contains only the outline of the image.
At a higher sampling rate, as shown in Fig. \ref{figure_video_sr0.2} and Fig. \ref{figure_hall_sr0.1},
the proposed models and competitive methods both recover the main structural information of the images,
but the proposed methods recover more texture and detail information.

\begin{table} [!t]
\caption{The averaged PSNR, SSIM, FSIM, ERGA and SAM of the recovered results on video "suzie" by Tmac, MF-TV, TNN, PSTNN and the proposed model-1, model-2 with different sampling rates. The best value is highlighted in bolder fonts.}
\label{table_video_suzie}
\resizebox{\textwidth}{42mm}{	
	\begin{tabular}{cccccccc}
		\hline \hline 				
		&&&SR =0.05&&&&	\\
		method&	Nosiy&	our model-2&	our model-1&	MF-TV&	Tmac&	PSTNN&	TNN\\		
		PSNR	&	7.259	&	\textbf{30.268}	&	26.663	&	13.801	&	23.385	&	17.447	&	22.005	\\
		SSIM	&	0.009	&	\textbf{0.85}	&	0.733	&	0.094	&	0.622	&	0.192	&	0.563	\\
		FSIM	&	0.454	&	\textbf{0.904}	&	0.852	&	0.42	&	0.792	&	0.59	&	0.776	\\
		ERGA	&	1057.282	&	\textbf{76.304}	&	115.628	&	501.117	&	167.927	&	327.678	&	194.844	\\
		MSAM	&	77.324	&	\textbf{3.258}	&	4.775	&	24.095	&	6.927	&	13.775	&	7.797	\\
		\hline
		&&&SR = 0.1&&&&\\
		method&	Nosiy&	our model-2&	our model-1&	MF-TV&	Tmac&	PSTNN&	TNN\\
		PSNR&	7.493&	\textbf{32.272}&	30.002&	22.356&	26.189&	26.647&	26.032\\
		SSIM&	0.014&	\textbf{0.887}&	0.832&	0.605&	0.74&	0.68&	0.692\\
		FSIM&	0.426&	\textbf{0.928}&	0.899&	0.758&	0.838&	0.843&	0.846\\
		ERGA&	1029.096&	\textbf{60.723}&	79.383&	196.059&	124.369	&117.104&	124.923\\
		MSAM&	71.725&	\textbf{2.678}&	3.385&	6.99&	5.423&	5.171&	5.405	\\	
		\hline	
		&&&SR = 0.2&&&&\\
		method&	Nosiy&	our model-2&	our model-1&	MF-TV&	Tmac&	PSTNN&	TNN\\
		PSNR&	8.005&	\textbf{34.492}&	33.745&	32.064&	27.274&	30.566&	30.561\\
		SSIM&	0.02&	\textbf{0.921}&	0.909&	0.872&	0.782&	0.829&	0.831\\
		FSIM&	0.391&	\textbf{0.95}&	0.943&	0.916&	0.853&	0.91&	0.911\\
		ERGA&	970.285&	\textbf{46.89}&	51.759&	66.692&	109.627&	75.472&	75.598\\
		MSAM&	63.522&	\textbf{2.142}&	2.329&	2.81&	4.812&	3.399&	3.395\\		
		\hline \hline
\end{tabular} }
\end{table}

\begin{table} [!t]
\centering	
\caption{The averaged PSNR, SSIM, FSIM, ERGA and SAM of the recovered results on video "hall" by Tmac, MF-TV, TNN, PSTNN and the proposed model-1, model-2 with different sampling rates. The best value is highlighted in bolder fonts.}
\label{table_video_hall}
\resizebox{\textwidth}{42mm}{
	\begin{tabular}{cccccccc}
		\hline \hline 				
		&&&SR =0.05&&&&	\\
		method&	Nosiy&	our model-2&	our model-1&	MF-TV&	Tmac&	PSTNN&	TNN\\
		PSNR	&	4.82	&	\textbf{29.571}	&	26.647	&	13.539	&	22.101	&	16.075	&	20.78	\\
		SSIM	&	0.007	&	\textbf{0.915}	&	0.862	&	0.412	&	0.675	&	0.36	&	0.636	\\
		FSIM	&	0.387	&	\textbf{0.935}	&	0.899	&	0.612	&	0.789	&	0.672	&	0.792	\\
		ERGA	&	1225.779	&	\textbf{73.007}	&	100.944	&	452.351	&	168.866	&	335.52	&	195.315	\\
		MSAM	&	77.299	&	\textbf{2.193}	&	2.727	&	12.865	&	3.818	&	8.64	&	4.299	\\
		\hline	
		&&&SR = 0.1&&&&\\		
		method	&	Nosiy	&	our model-2	&	our model-1	&	MF-TV	&	Tmac	&	PSTNN	&	TNN	\\
		PSNR	&	5.055	&	\textbf{32.103}	&	30.241	&	24.855	&	26.936	&	29.014	&	28.433	\\
		SSIM	&	0.013	&	\textbf{0.936}	&	0.918	&	0.829	&	0.854	&	0.892	&	0.905	\\
		FSIM	&	0.393	&	\textbf{0.953}	&	0.939	&	0.873	&	0.888	&	0.934	&	0.936	\\
		ERGA	&	1193.075	&	\textbf{55.089}	&	67.967	&	131.422	&	97.185	&	77.395	&	82.259	\\
		MSAM	&	71.7	&\textbf{	1.824}	&	2.11	&	3.669	&	2.404	&	2.417	&	2.46	\\	
		\hline
		&&&SR = 0.2&&&&\\
		method	&	Nosiy	&	our model-2	&	our model-1	&	MF-TV	&	Tmac	&	PSTNN	&	TNN	\\
		PSNR	&	5.567	&	\textbf{34.045}	&	33.647	&	33.006	&	27.648	&	33.629	&	33.691	\\
		SSIM	&	0.025	&	\textbf{0.953}	&	0.952	&	0.94	&	0.869	&	0.961	&	0.962	\\
		FSIM	&	0.403	&	\textbf{0.965}	&	0.964	&	0.954	&	0.897	&	0.973	&	0.974	\\
		ERGA	&	1124.737	&	\textbf{43.939}	&	46.002	&	50.971	&	89.271	&	46.123	&	45.851	\\
		MSAM	&	63.507	&	\textbf{1.546}	&	1.584	&	1.779	&	2.226	&	1.584	&	1.565	\\		
		\hline \hline
\end{tabular}}
\end{table}

\begin{figure*}[!t]	
\centering
\includegraphics[width=0.3\linewidth]{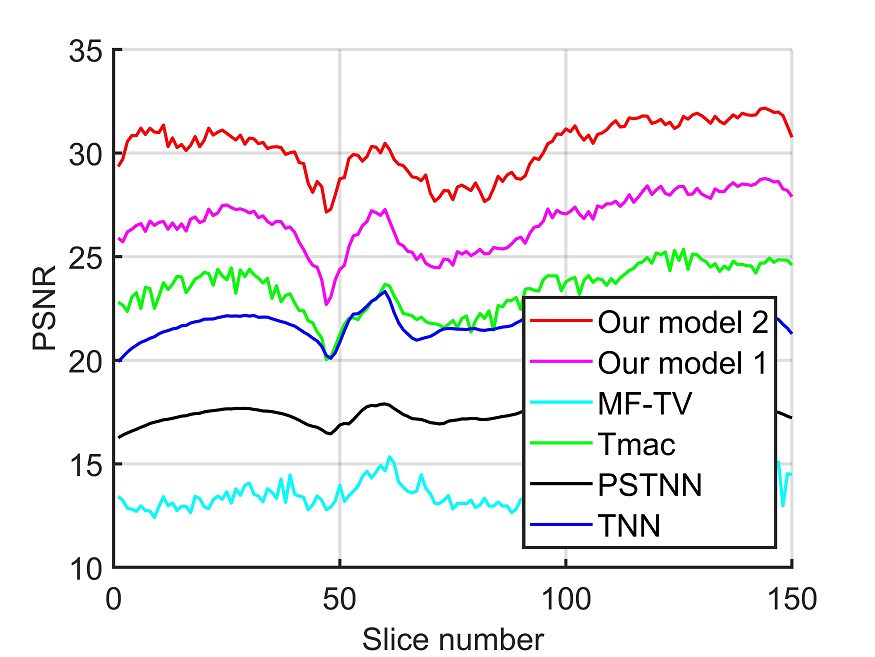}
\hfil
\includegraphics[width=0.3\linewidth]{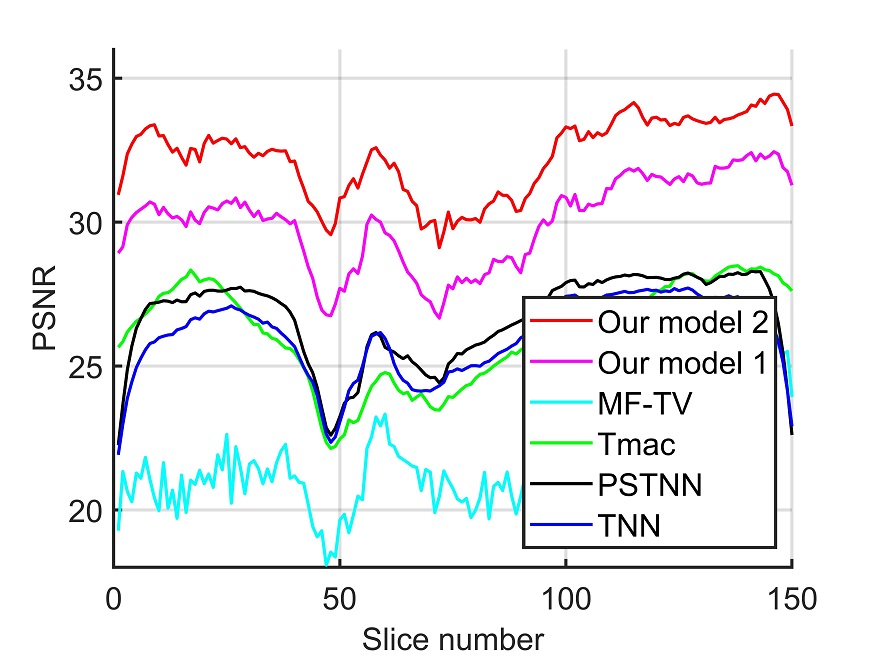}
\hfil
\includegraphics[width=0.3\linewidth]{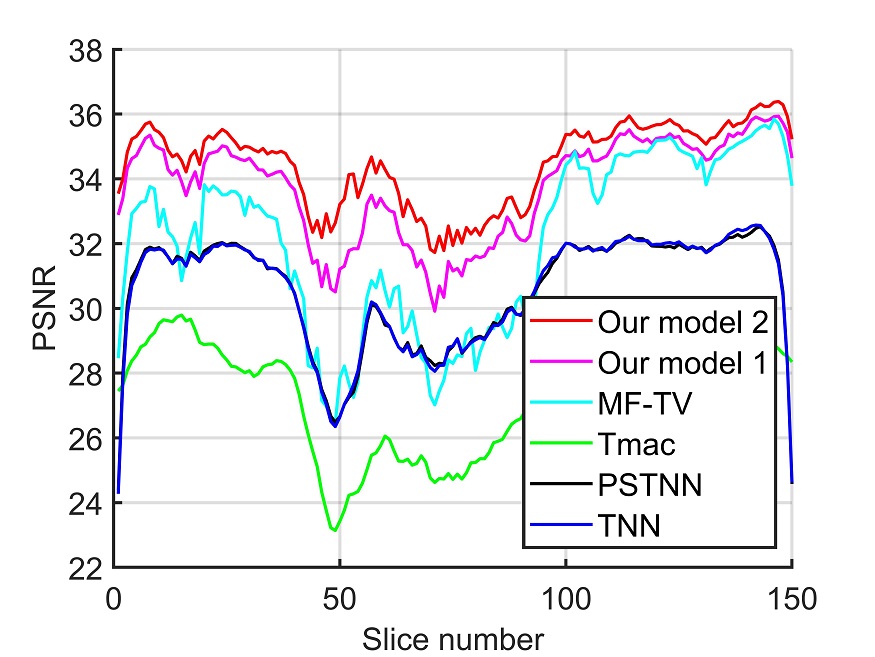}
\hfil
\includegraphics[width=0.3\linewidth]{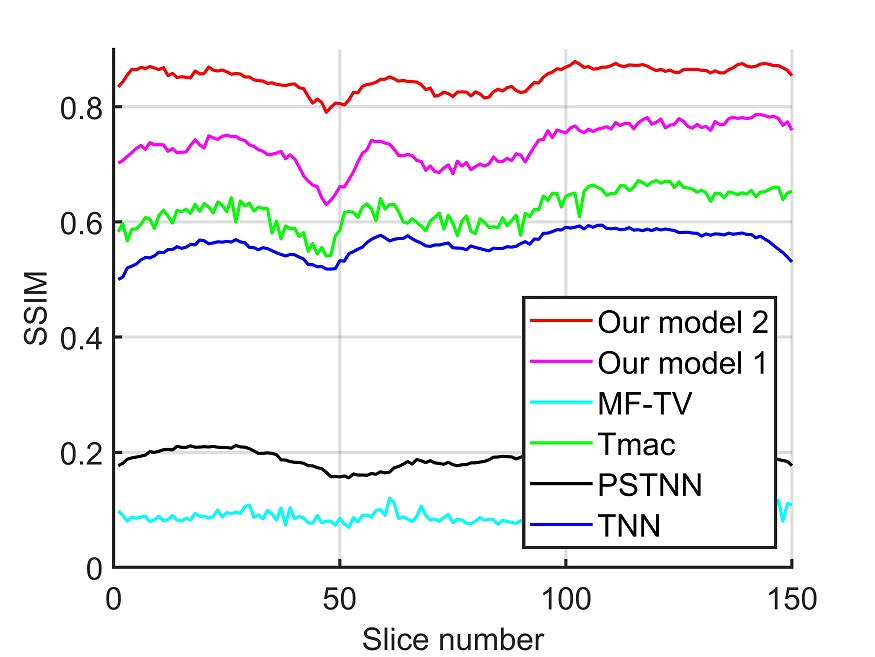}
\hfil
\includegraphics[width=0.3\linewidth]{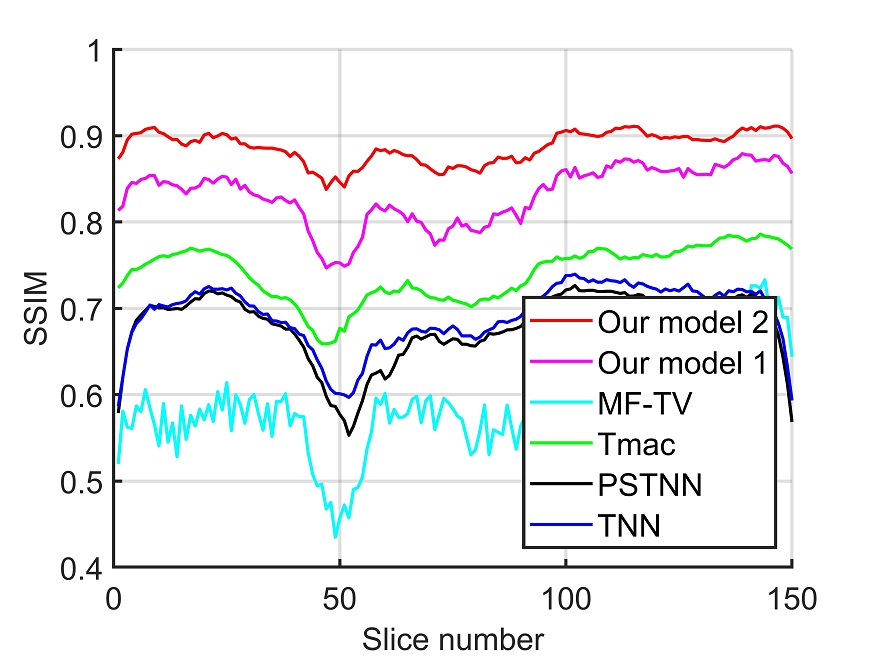}
\hfil
\includegraphics[width=0.3\linewidth]{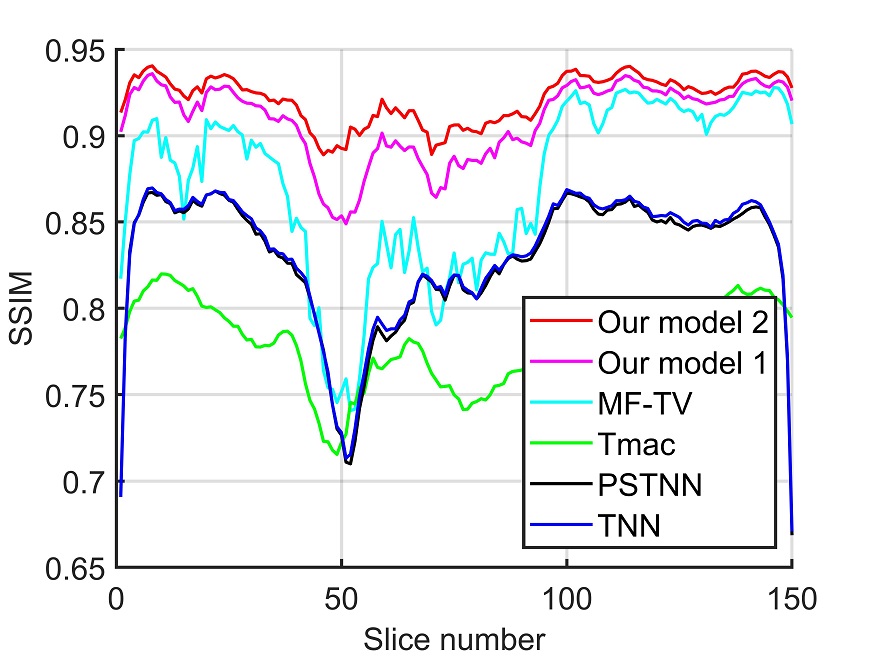}	
\hfil				
\subfloat[SR = 0.05]{\includegraphics[width=0.3\linewidth]{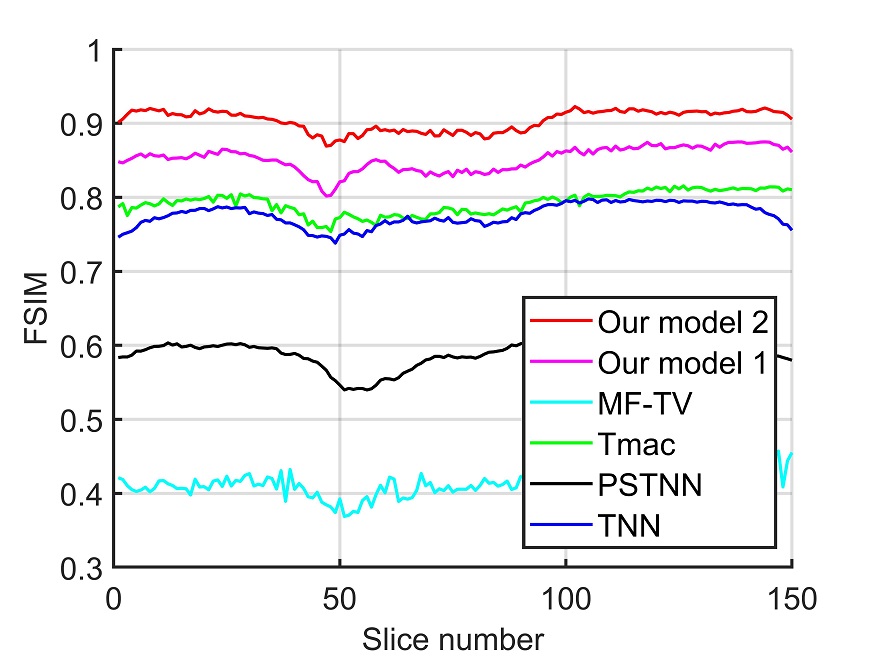}}%
\hfil
\subfloat[SR = 0.1]{\includegraphics[width=0.3\linewidth]{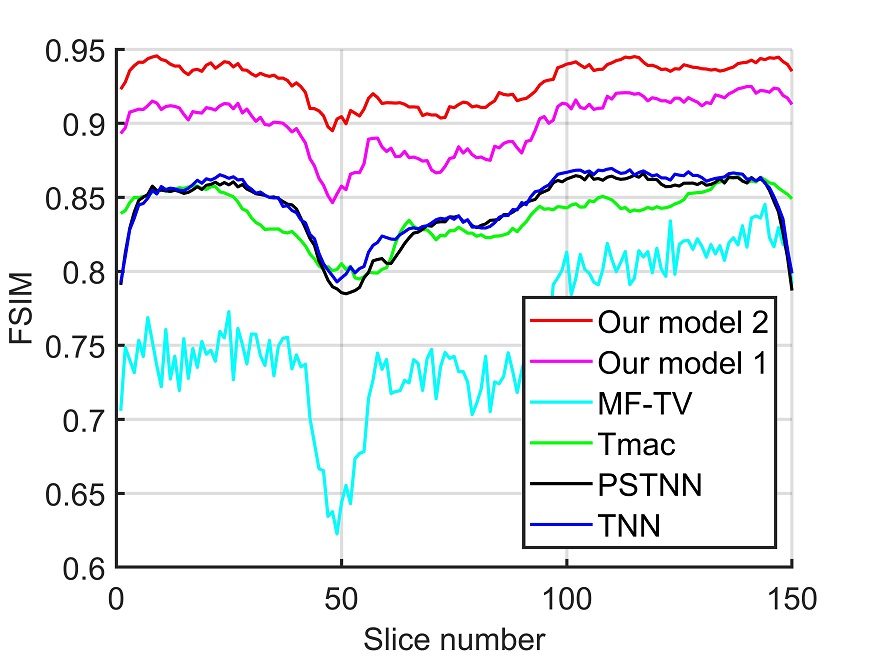}}%
\hfil
\subfloat[SR = 0.2]{\includegraphics[width=0.3\linewidth]{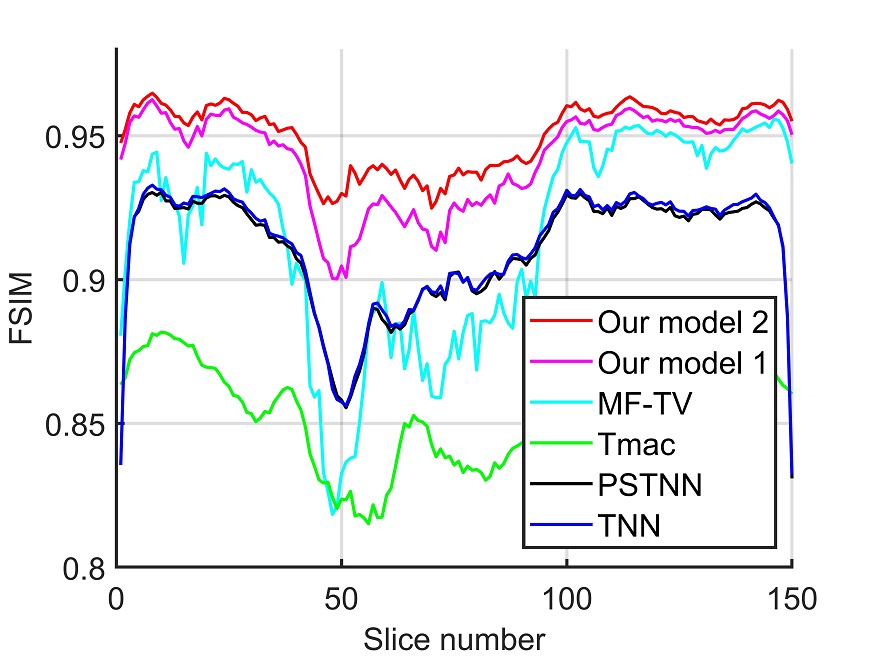}}%
\caption{The PSNR, SSIM and FSIM of the recovered video "suzie" by MF-TV, Tmac, TNN, PSTNN and our model-1 and model-2 for all slices, respectively.}
\label{PSNR and SSIM of video}
\end{figure*}

\subsection{MRI}

In this part, to further verify the versatility of our models for different datasets,
the proposed models are applied to MRI dataset, i.e., the cubical MRI data\footnote{http://brainweb.bic.mni.mcgill.ca/brainweb/selection$\_$normal.html}.
The size of the dataset is 150 $\times$ 150 $\times$ 181.
The sampling rates are set as 5\%, 10\%, 20\% and 30\%.

\begin{figure*}[!t]	
\centering			
\subfloat[Original]{\includegraphics[width=0.23\linewidth]{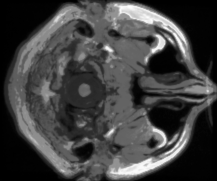}}%
\hfil	
\subfloat[90\% Masked]{\includegraphics[width=0.23\linewidth]{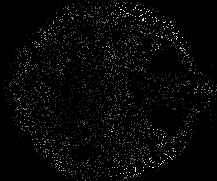}}%
\hfil					
\subfloat[our model-2]{\includegraphics[width=0.23\linewidth]{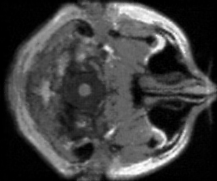}}%
\hfil
\subfloat[our model-1]{\includegraphics[width=0.23\linewidth]{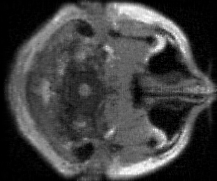}}%
\hfil
\subfloat[MF-TV]{\includegraphics[width=0.23\linewidth]{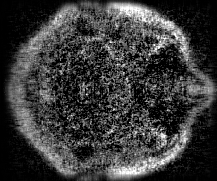}}%
\hfil					
\subfloat[Tmac]{\includegraphics[width=0.23\linewidth]{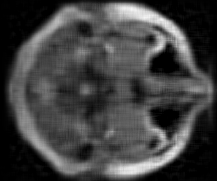}}%
\hfil
\subfloat[PSTNN]{\includegraphics[width=0.23\linewidth]{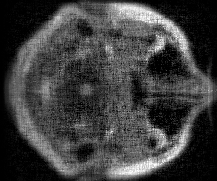}}%
\hfil
\subfloat[TNN]{\includegraphics[width=0.23\linewidth]{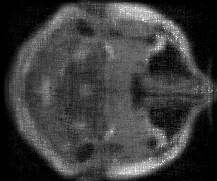}}%
\caption{One slice of the recovered MRI by our model-1 and model-2, MF-TV, Tmac, PSTNN and TNN.  The sampling rate is 10\%.}
\label{figure_MR_sr0.1_1}
\end{figure*}

\begin{figure*}[!t]		
\centering			
\subfloat[Original]{\includegraphics[width=0.23\linewidth]{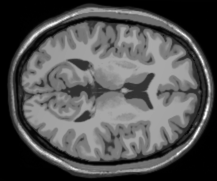}}%
\hfil
\subfloat[90\% Masked]{\includegraphics[width=0.23\linewidth]{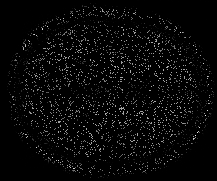}}%
\hfil					
\subfloat[our model-2]{\includegraphics[width=0.23\linewidth]{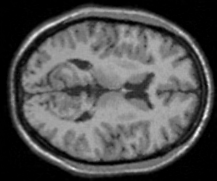}}%
\hfil
\subfloat[our model-1]{\includegraphics[width=0.23\linewidth]{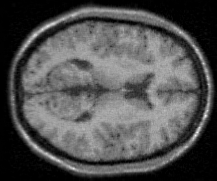}}%
\hfil
\subfloat[MF-TV]{\includegraphics[width=0.23\linewidth]{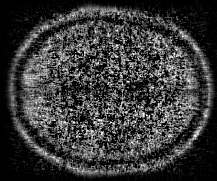}}%
\hfil					
\subfloat[Tmac]{\includegraphics[width=0.23\linewidth]{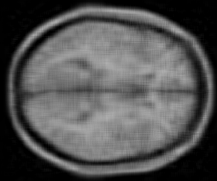}}%
\hfil
\subfloat[PSTNN]{\includegraphics[width=0.23\linewidth]{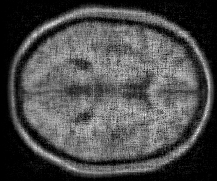}}%
\hfil
\subfloat[TNN]{\includegraphics[width=0.23\linewidth]{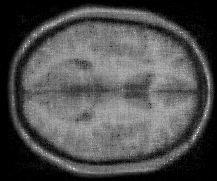}}%
\caption{One slice of the recovered MRI by our model-1 and model-2, MF-TV, Tmac, PSTNN and TNN.  The sampling rate is 10\%.}
\label{figure_MR_sr0.1_2}
\end{figure*}

\begin{figure*}[!t]	
\centering			
\subfloat[Original]{\includegraphics[width=0.23\linewidth]{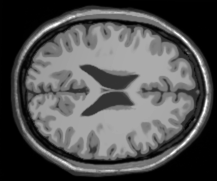}}%
\hfil
\subfloat[90\% Masked]{\includegraphics[width=0.23\linewidth]{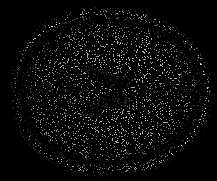}}%
\hfil					
\subfloat[our model-2]{\includegraphics[width=0.23\linewidth]{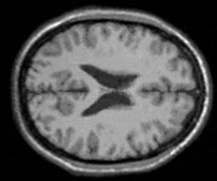}}%
\hfil
\subfloat[our model-1]{\includegraphics[width=0.23\linewidth]{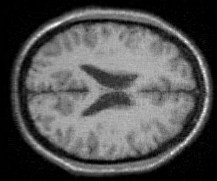}}%
\hfil
\subfloat[MF-TV]{\includegraphics[width=0.23\linewidth]{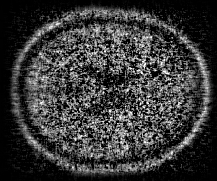}}%
\hfil					
\subfloat[Tmac]{\includegraphics[width=0.23\linewidth]{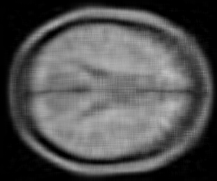}}%
\hfil
\subfloat[PSTNN]{\includegraphics[width=0.23\linewidth]{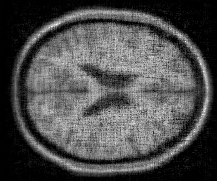}}%
\hfil
\subfloat[TNN]{\includegraphics[width=0.23\linewidth]{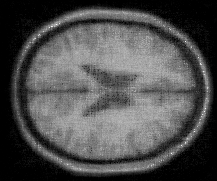}}%
\caption{One slice of the recovered MRI by our model-1 and model-2, MF-TV, Tmac, PSTNN and TNN.  The sampling rate is 10\%.}
\label{figure_MR_sr0.1_3}
\end{figure*}

\begin{figure*}[!t]	
\centering			
\subfloat[Original]{\includegraphics[width=0.23\linewidth]{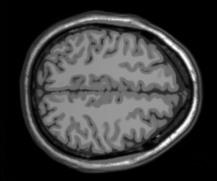}}%
\hfil	
\subfloat[90\% Masked]{\includegraphics[width=0.23\linewidth]{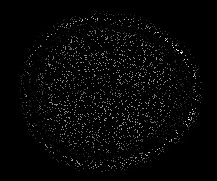}}%
\hfil					
\subfloat[our model-2]{\includegraphics[width=0.23\linewidth]{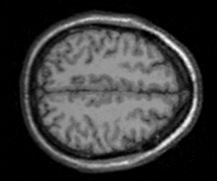}}%
\hfil
\subfloat[our model-1]{\includegraphics[width=0.23\linewidth]{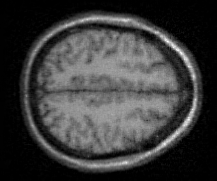}}%
\hfil
\subfloat[MF-TV]{\includegraphics[width=0.23\linewidth]{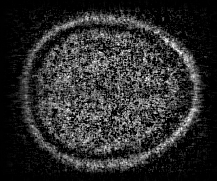}}%
\hfil					
\subfloat[Tmac]{\includegraphics[width=0.23\linewidth]{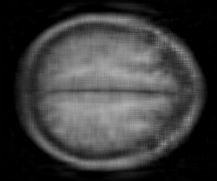}}%
\hfil
\subfloat[PSTNN]{\includegraphics[width=0.23\linewidth]{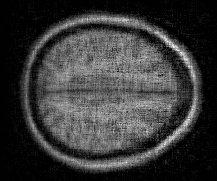}}%
\hfil
\subfloat[TNN]{\includegraphics[width=0.23\linewidth]{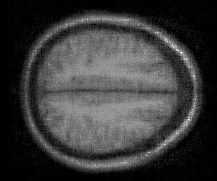}}%
\caption{One slice of the recovered MRI by our model-1 and model-2, MF-TV, Tmac, PSTNN and TNN.  The sampling rate is 10\%.}
\label{figure_MR_sr0.1_4}
\end{figure*}

\begin{figure*}[!t]	
\centering
\includegraphics[width=0.3\linewidth]{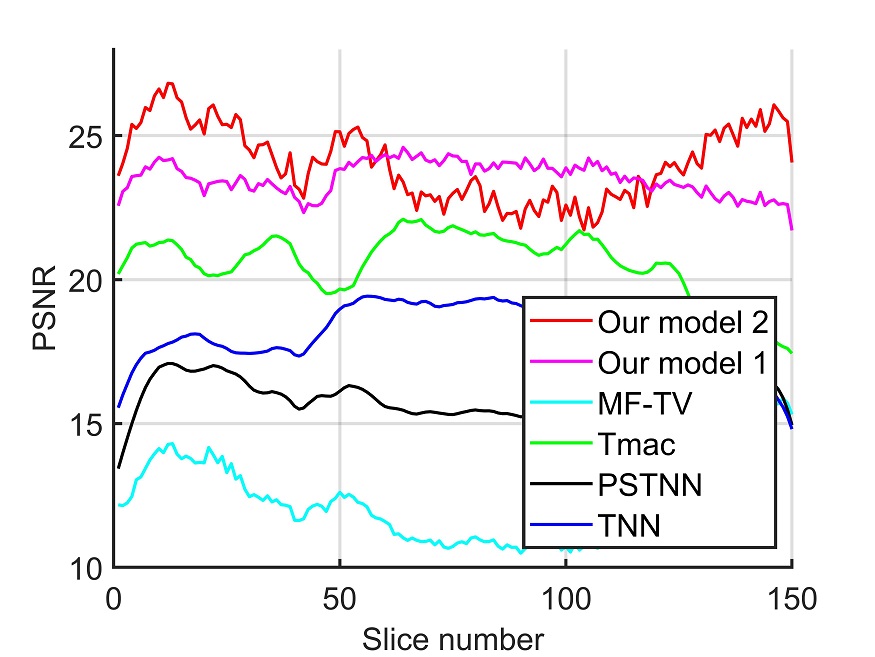}
\hfil	
\includegraphics[width=0.3\linewidth]{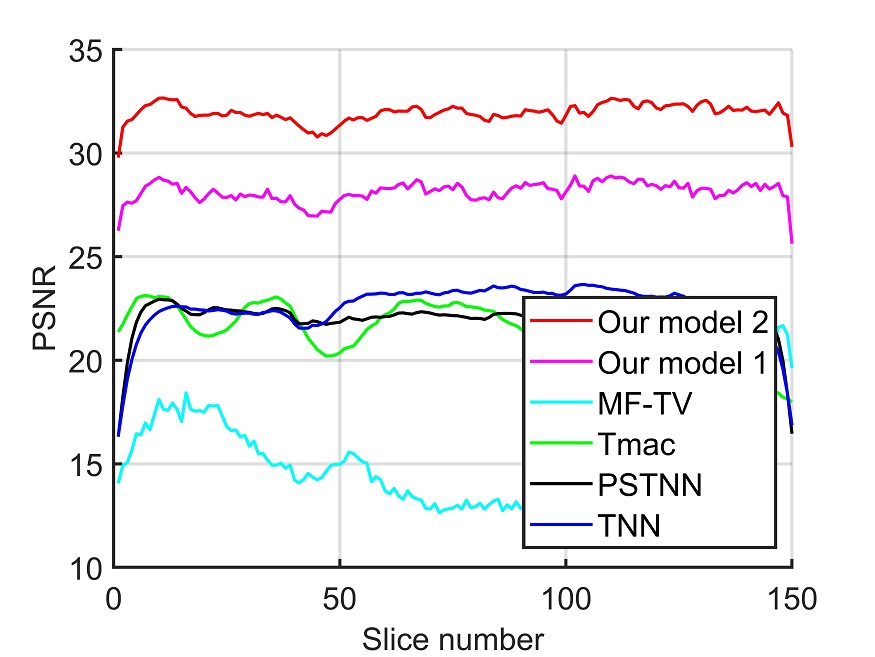}
\hfil
\includegraphics[width=0.3\linewidth]{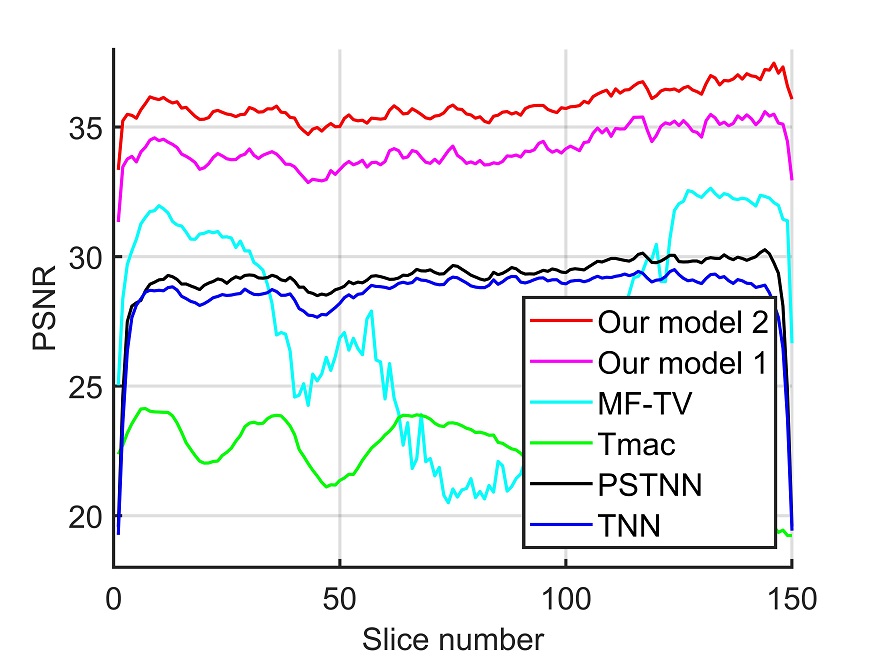}
\hfil
\includegraphics[width=0.3\linewidth]{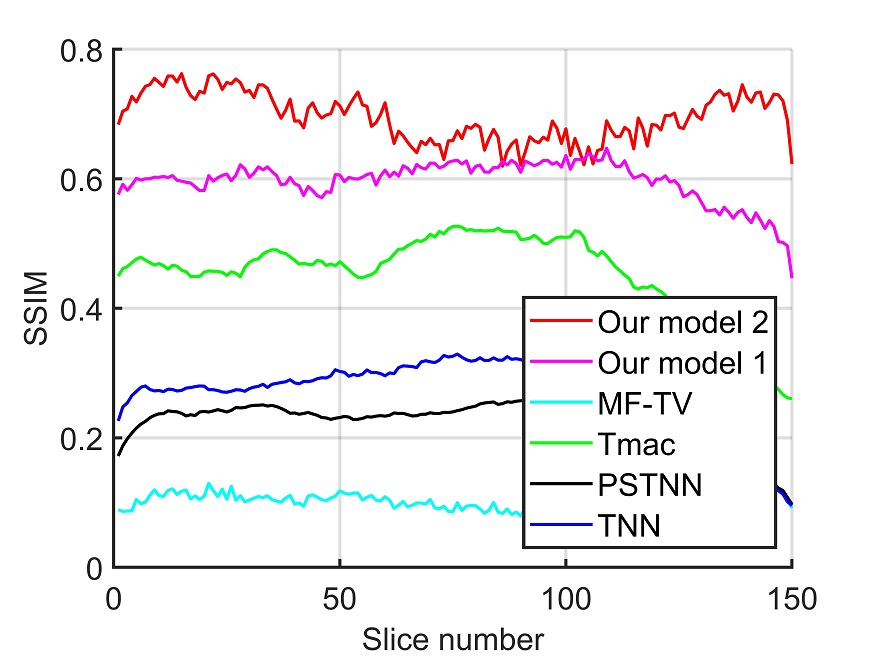}
\hfil
\includegraphics[width=0.3\linewidth]{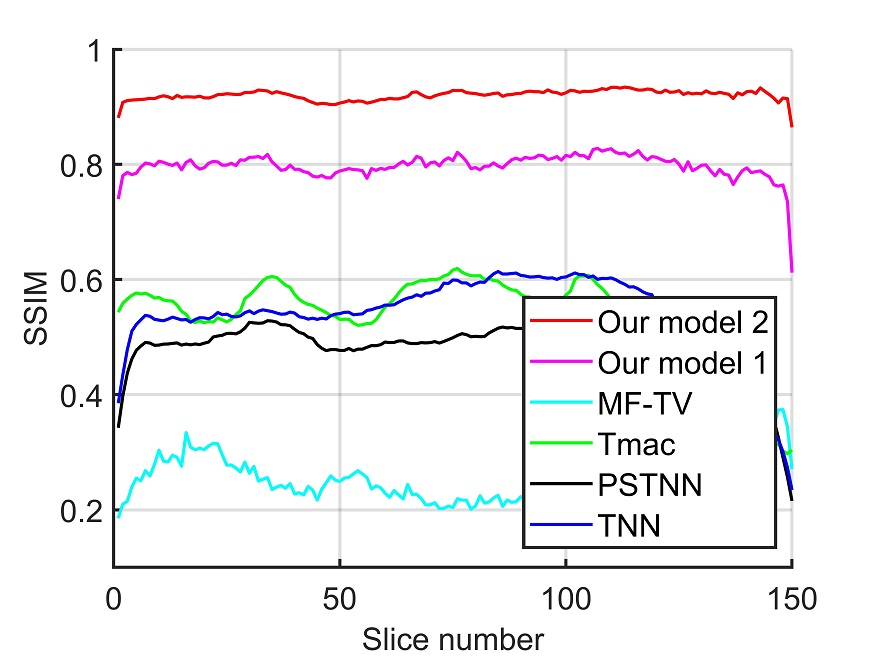}
\hfil
\includegraphics[width=0.3\linewidth]{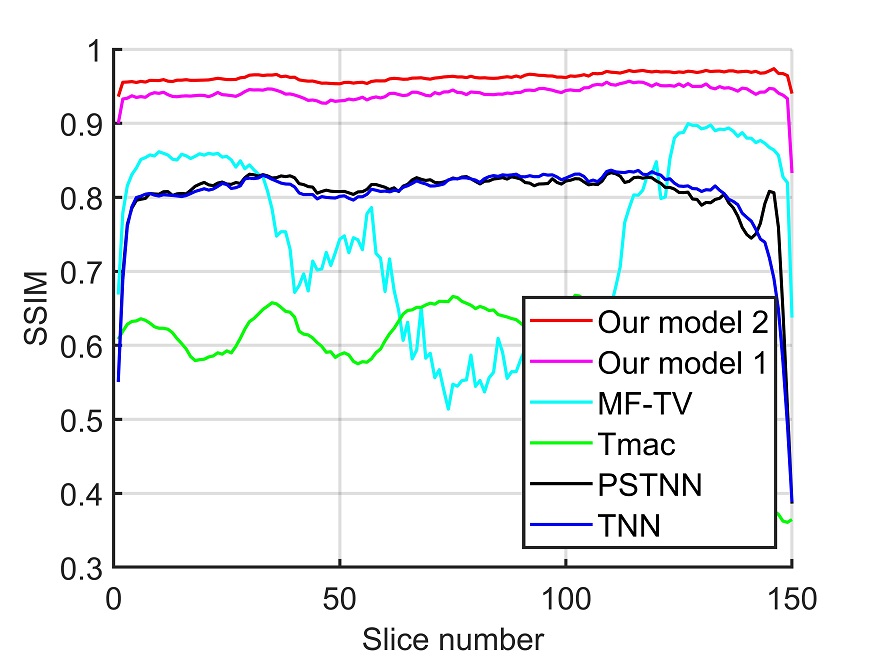}
\hfil
\subfloat[SR = 0.05]{\includegraphics[width=0.3\linewidth]{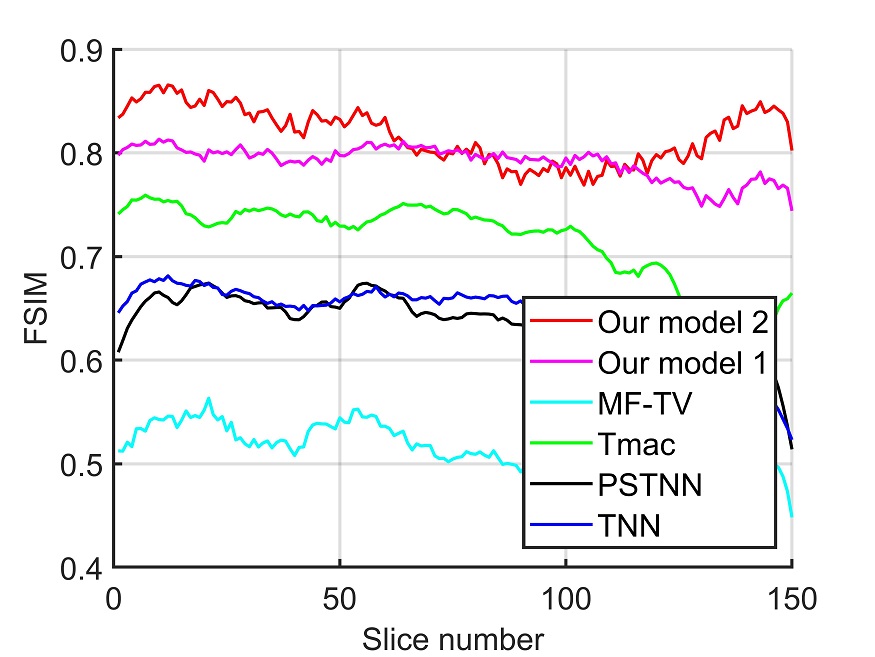}}%
\hfil					
\subfloat[SR = 0.1]{\includegraphics[width=0.3\linewidth]{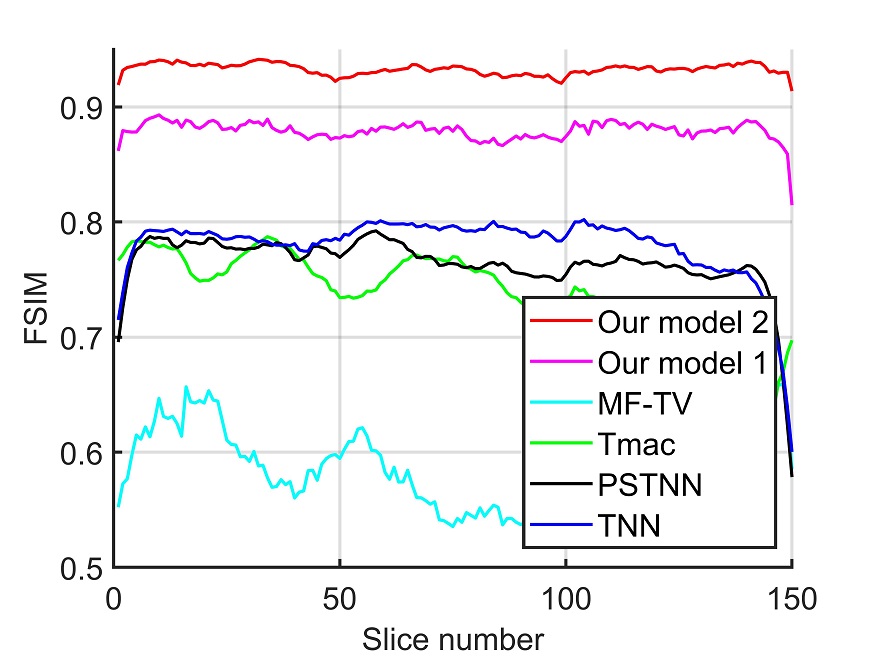}}%
\hfil
\subfloat[SR = 0.2]{\includegraphics[width=0.3\linewidth]{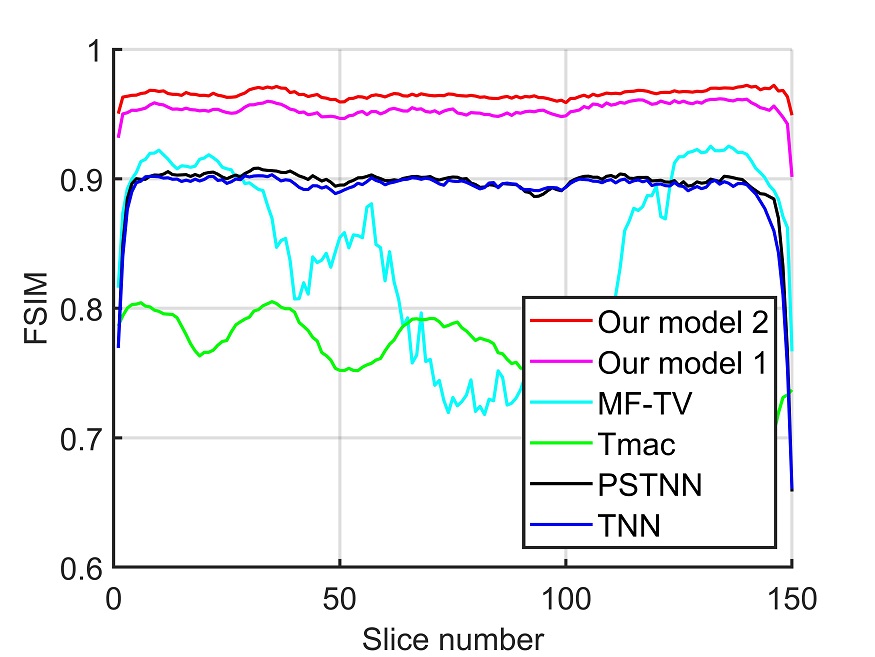}}%
\caption{The PSNR, SSIM and FSIM of the recovered MRI by MF-TV, Tmac, TNN, PSTNN and our model-1 and model-2 for all slices, respectively.}
\label{PSNR and SSIM of MRI}
\end{figure*}

Table \ref{table_MRI} summarizes the PQIs of the recovered results at the four sampling rates in the MRI dataset.
It can be clearly found that our proposed models achieve higher PQIs than the comparative models.
And the same advantage of our models can also be seen in Fig. \ref{PSNR and SSIM of MRI},
which illustrates the PSNR, SSIM and FSIM values slice by slice in all sampling rates.
For visual comparison, at a sampling rate of 0.1,
Fig. \ref{figure_MR_sr0.1_1},
Fig. \ref{figure_MR_sr0.1_2},
Fig. \ref{figure_MR_sr0.1_3} and
Fig. \ref{figure_MR_sr0.1_4} show the gray-scale images of the original MRI data, the sampled data, and the different recovered results.
It can be seen that our models can better retain the local details and texture information of the images, and effectively restore the main structure of the image.
Therefore, one can see that the recovered data obtained by our models has the best visual evaluation.

\begin{table}

\centering
\caption{The averaged PSNR, SSIM, FSIM, ERGA and SAM of the recovered results on MRI by Tmac, MF-TV, TNN, PSTNN and the proposed model-1, model-2 with different sampling rates. The best value is highlighted in bolder fonts.}
\label{table_MRI}
\resizebox{\textwidth}{53mm}{
	\begin{tabular}{ccccccccc}
		\hline \hline
		&&&SR =0.05 &&&&	\\
		method&	Nosiy&	our model-2&	our model-1&	MF-TV&	Tmac&	PSTNN&	TNN\\
		PSNR	&	10.258	&	\textbf{24.048}	&	23.54	&	12.332	&	20.51	&	15.859	&	18.218	\\
		SSIM	&	0.228	&	\textbf{0.696}	&	0.597	&	0.099	&	0.45	&	0.224	&	0.27	\\
		FSIM	&	0.473	&	\textbf{0.817}	&	0.791	&	0.52	&	0.711	&	0.642	&	0.646	\\
		ERGA	&	1030.203	&	\textbf{212.967}	&	230.079	&	814.747	&	339.385	&	545.77	&	434.774	\\
		MSAM	&	76.54	&	\textbf{20.912}	&	22.626	&	55.603	&	31.367	&	36.355	&	31.11	\\
		\hline
		&&&SR = 0.1&&&&\\				
		method	&	Nosiy	&	our model-2	&	our model-1	&	MF-TV	&	Tmac	&	PSTNN	&	TNN	\\
		PSNR	&	10.492	&	\textbf{31.9}	&	28.085	&	15.406	&	21.411	&	22.061	&	22.535	\\
		SSIM	&	0.241	&	\textbf{0.919}	&	0.798	&	0.25	&	0.531	&	0.482	&	0.536	\\
		FSIM	&	0.511	&	\textbf{0.932}	&	0.879	&	0.587	&	0.732	&	0.764	&	0.78	\\
		ERGA	&	1002.8	&	\textbf{86.415}	&	134.58	&	584.827	&	308.655	&	275.473	&	266.753	\\
		MSAM	&	70.986	&	\textbf{14.285}	&	18.022	&	41.826	&	29.345	&	24.585	&	24.6	\\
		\hline
		&&&SR = 0.2&&&&\\
		method	&	Nosiy	&	our model-2	&	our model-1	&	MF-TV	&	Tmac	&	PSTNN	&	TNN	\\
		PSNR	&	11.003	&	\textbf{35.842}	&	34.166	&	27.062	&	22.33	&	29.152	&	28.571	\\
		SSIM	&	0.271	&	\textbf{0.963}	&	0.941	&	0.737	&	0.586	&	0.804	&	0.802	\\
		FSIM	&	0.564	&	\textbf{0.965}	&	0.954	&	0.84	&	0.754	&	0.895	&	0.891	\\
		ERGA	&	945.583	&	\textbf{54.522}	&	66.369	&	173.636	&	276.269	&	127.133	&	136.182	\\
		MSAM	&	62.887	&	\textbf{11.855}	&	13.38	&	21.792	&	27.267	&	17.513	&	17.855	\\
		\hline
		&&&SR = 0.3&&&&\\
		method	&	Nosiy	&	our model-2	&	our model-1	&	MF-TV	&	Tmac	&	PSTNN	&	TNN	\\
		PSNR	&	11.582	&	\textbf{37.991}	&	37.301	&	36.355	&	23.077	&	32.608	&	32.481	\\
		SSIM	&	0.303	&	\textbf{0.977}	&	0.971	&	0.954	&	0.625	&	0.895	&	0.89	\\
		FSIM	&	0.597	&	\textbf{0.978}	&	0.975	&	0.962	&	0.773	&	0.939	&	0.939	\\
		ERGA	&	884.608	&	\textbf{42.534}	&	46.165	&	52.449	&	252.057	&	85.845	&	87.312	\\
		MSAM	&	56.216	&	\textbf{10.284}	&	11.019	&	13.849	&	25.722	&	14.858	&	14.879	\\
		\hline \hline
\end{tabular} }
\end{table}

\subsection{Hyperspectral image}

In this subsection, we select two HSI data to apply simulated experiments.
The first dataset is the Pavia City Centre\footnote{http://www.ehu.es/ccwintco/index.php/Hyperspectral$\_$Remote$\_$Sensing$\_$Scenes}
which was filmed by the reflection optical system imaging spectrometer (ROSIS-03).
Its size is $1096\times1096$, with a total of 102 bands.
Because some of the bands in the Pavia City Centre dataset are heavily polluted by noise, they can not be used as a reference for restoration results.
Therefore, this part of the heavily polluted data has been removed.
Due to space limitations, we select data with a spatial size of $200\times200$ and a total of 80 bands for simulated experiments in this part.
The second dataset is the Airborne Visible/Infrared Imaging Spectrometer (AVIRIS) Cuprite data\footnote{http://aviris.jpl.nasa.gov/html/aviris.freedata.html}.
Its size is 150 $\times$ 150 $\times$ 210.
The sampling rate is set to 0.025, 0.05 and 0.1.

\begin{figure*}[t]	
\centering			
\subfloat[Original]{\includegraphics[width=0.23\linewidth]{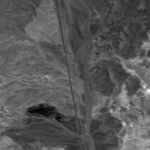}}%
\hfil	
\subfloat[95\% Masked]{\includegraphics[width=0.23\linewidth]{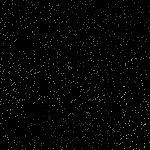}}%
\hfil					
\subfloat[our model-2]{\includegraphics[width=0.23\linewidth]{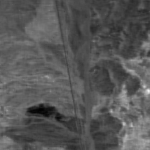}}%
\hfil
\subfloat[our model-1]{\includegraphics[width=0.23\linewidth]{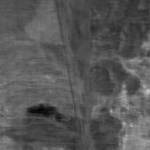}}%
\hfil
\subfloat[MF-TV]{\includegraphics[width=0.23\linewidth]{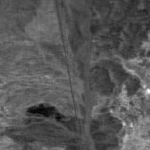}}%
\hfil					
\subfloat[Tmac]{\includegraphics[width=0.23\linewidth]{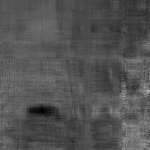}}%
\hfil
\subfloat[PSTNN]{\includegraphics[width=0.23\linewidth]{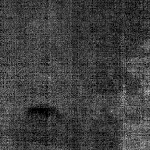}}%
\hfil
\subfloat[TNN]{\includegraphics[width=0.23\linewidth]{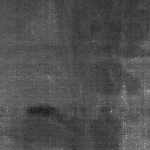}}%
\caption{One slice of the recovered HSI "Cuprite" by our model-1 and model-2, MF-TV, Tmac, PSTNN and TNN.  The sampling rate is 5\%.}
\label{figure_HSI_sr0.05}
\end{figure*}

\begin{figure*}[t]	
\centering
\subfloat[PSNR]{\includegraphics[width=0.3\linewidth]{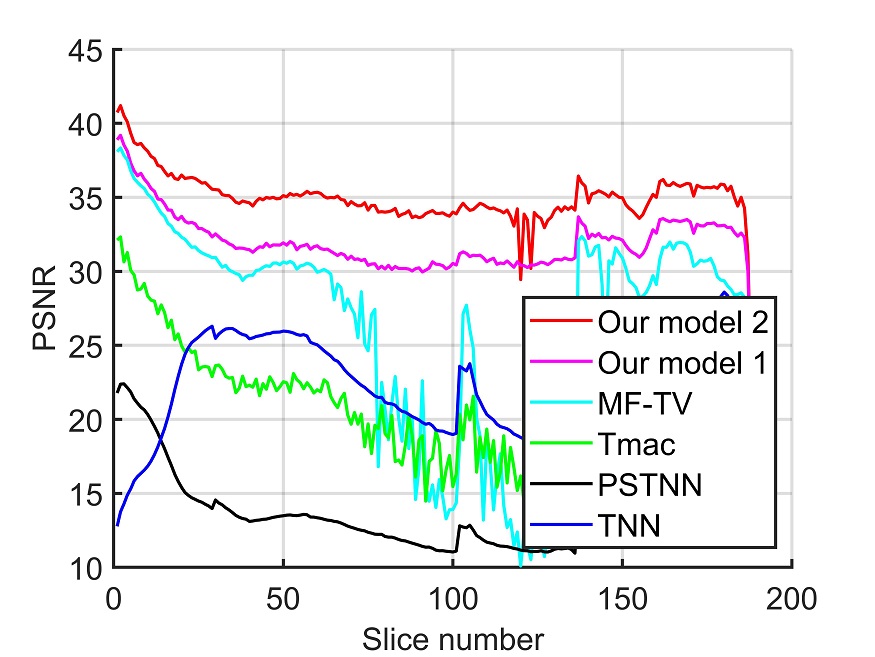}}%
\hfil					
\subfloat[SSIM]{\includegraphics[width=0.3\linewidth]{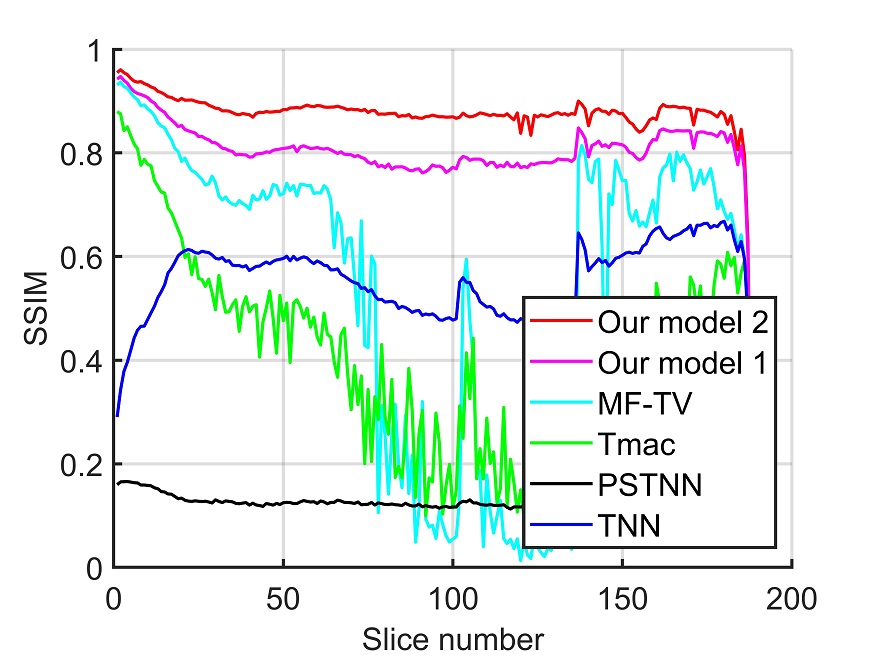}}%
\hfil
\subfloat[FSIM]{\includegraphics[width=0.3\linewidth]{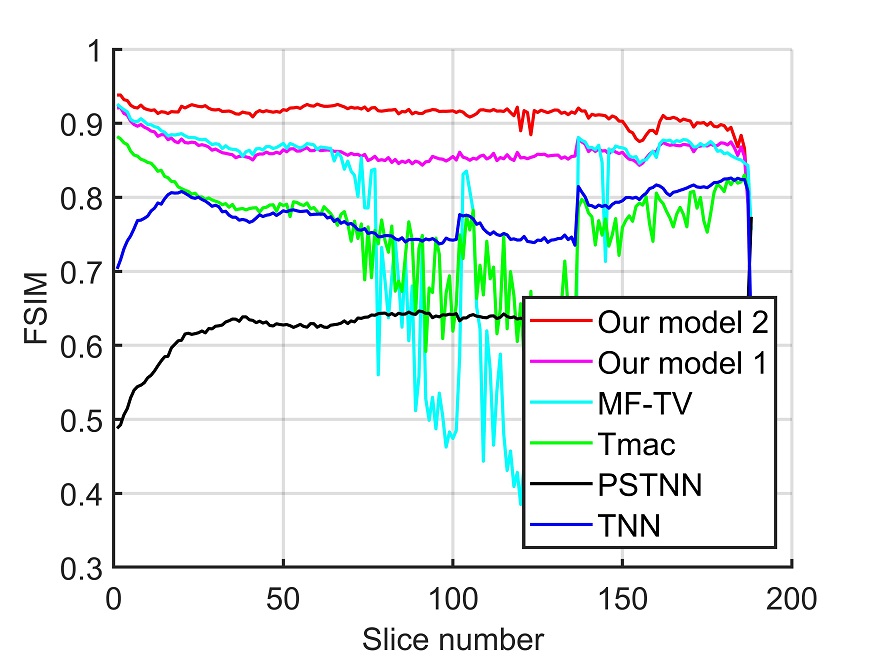}}%
\hfil	
\subfloat[PSNR]{\includegraphics[width=0.3\linewidth]{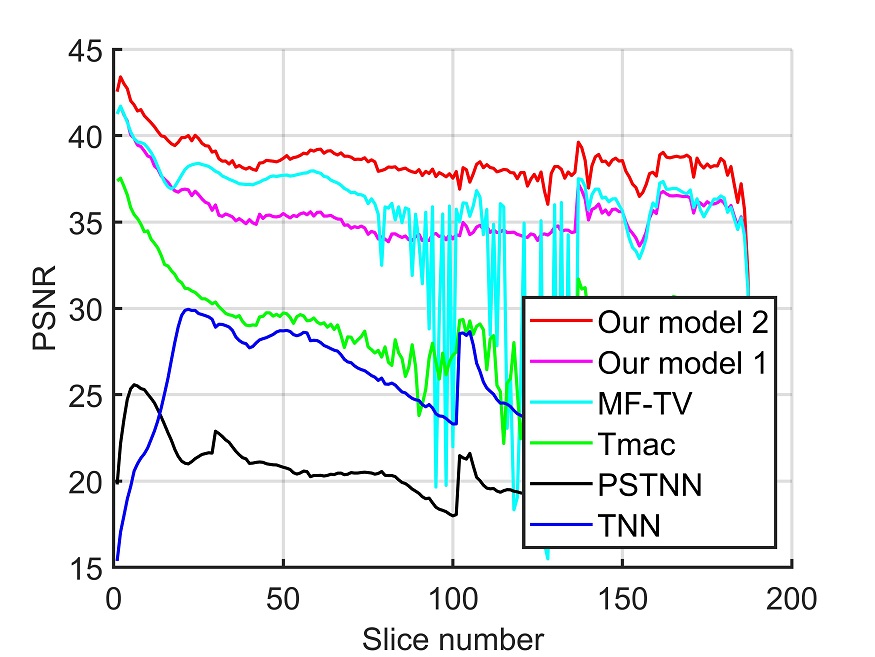}}%
\hfil					
\subfloat[SSIM]{\includegraphics[width=0.3\linewidth]{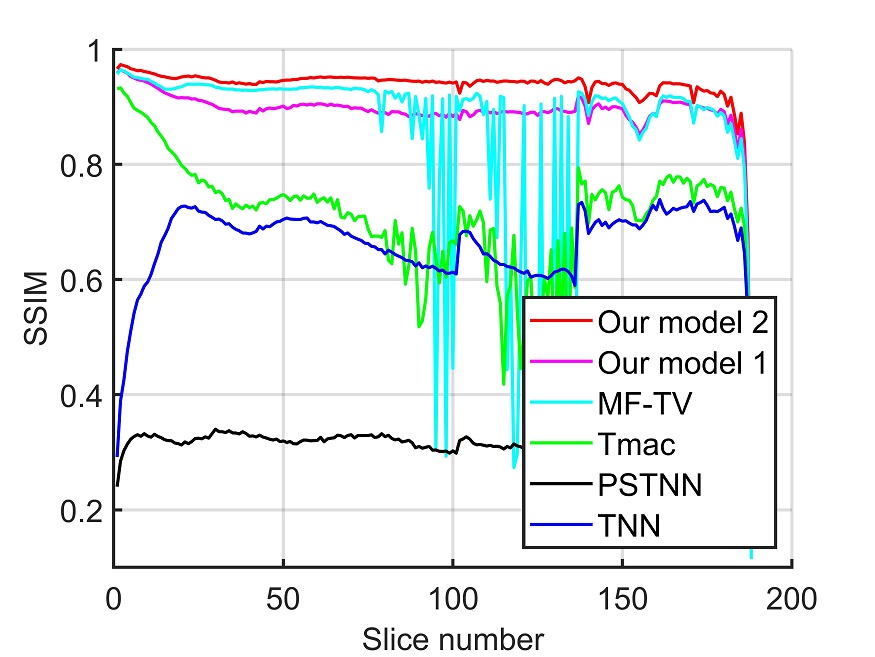}}%
\hfil
\subfloat[FSIM]{\includegraphics[width=0.3\linewidth]{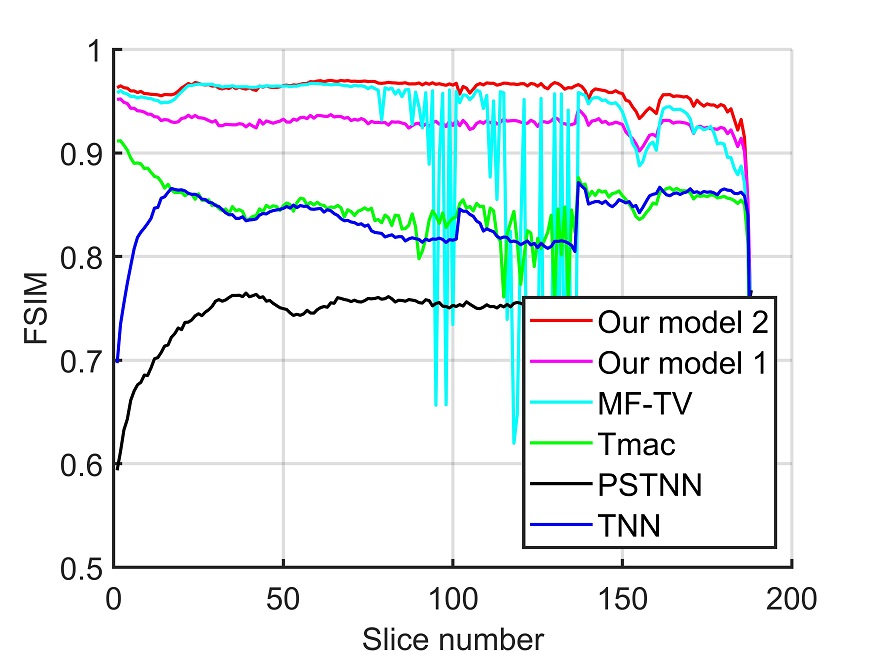}}%
\caption{The PSNR, SSIM and FSIM of the recovered HSI "Cuprite" by MF-TV, Tmac, TNN, PSTNN and our model-1 and model-2 for all slices, respectively.(a)-(c): 97.5\% entries missing, (d)-(f): 95\% entries missing.}
\label{PSNR and SSIM of HSI}
\end{figure*}

\begin{figure*}[t]	
\centering			
\subfloat[Original]{\includegraphics[width=0.23\linewidth]{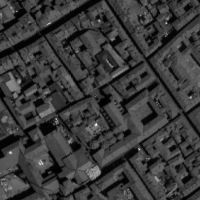}}%
\hfil	
\subfloat[97.5\% Masked]{\includegraphics[width=0.23\linewidth]{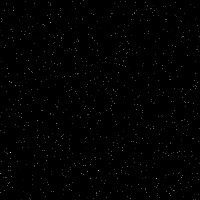}}%
\hfil					
\subfloat[our-model 2]{\includegraphics[width=0.23\linewidth]{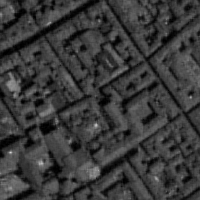}}%
\hfil
\subfloat[our model-1]{\includegraphics[width=0.23\linewidth]{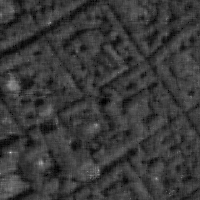}}%
\hfil
\subfloat[MF-TV]{\includegraphics[width=0.23\linewidth]{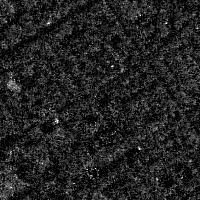}}%
\hfil					
\subfloat[Tmac]{\includegraphics[width=0.23\linewidth]{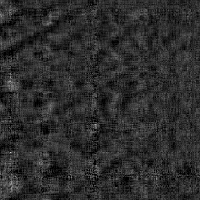}}%
\hfil
\subfloat[PSTNN]{\includegraphics[width=0.23\linewidth]{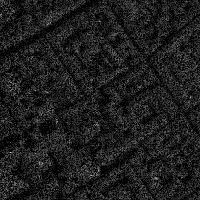}}%
\hfil
\subfloat[TNN]{\includegraphics[width=0.23\linewidth]{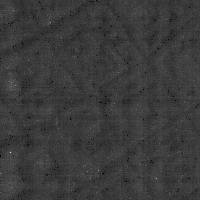}}%
\caption{One slice of the recovered HSI "Pavia" by our model-1 and model-2, MF-TV, Tmac, PSTNN and TNN.  The sampling rate is 2.5\%.}
\label{figure_Pavia_sr0.025}
\end{figure*}

\begin{figure*}[t]	
\centering			
\subfloat[Original]{\includegraphics[width=0.23\linewidth]{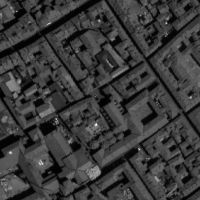}}%
\hfil	
\subfloat[95\% Masked]{\includegraphics[width=0.23\linewidth]{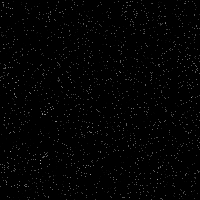}}%
\hfil					
\subfloat[our model-2]{\includegraphics[width=0.23\linewidth]{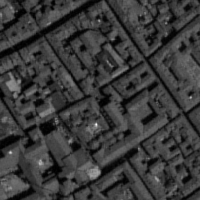}}%
\hfil
\subfloat[our model-1]{\includegraphics[width=0.23\linewidth]{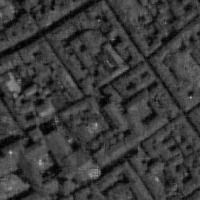}}%
\hfil
\subfloat[MF-TV]{\includegraphics[width=0.23\linewidth]{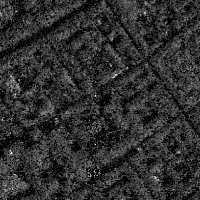}}%
\hfil					
\subfloat[Tmac]{\includegraphics[width=0.23\linewidth]{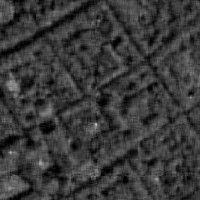}}%
\hfil
\subfloat[PSTNN]{\includegraphics[width=0.23\linewidth]{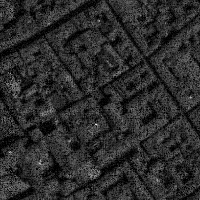}}%
\hfil
\subfloat[TNN]{\includegraphics[width=0.23\linewidth]{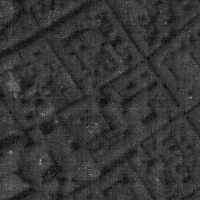}}%
\caption{One slice of the recovered HSI "Pavia" by our model-1 and model-2, MF-TV, Tmac, PSTNN and TNN.  The sampling rate is 5\%.}
\label{figure_Pavia_sr0.05}
\end{figure*}

\begin{figure*}[t]	
\centering
\includegraphics[width=0.3\linewidth]{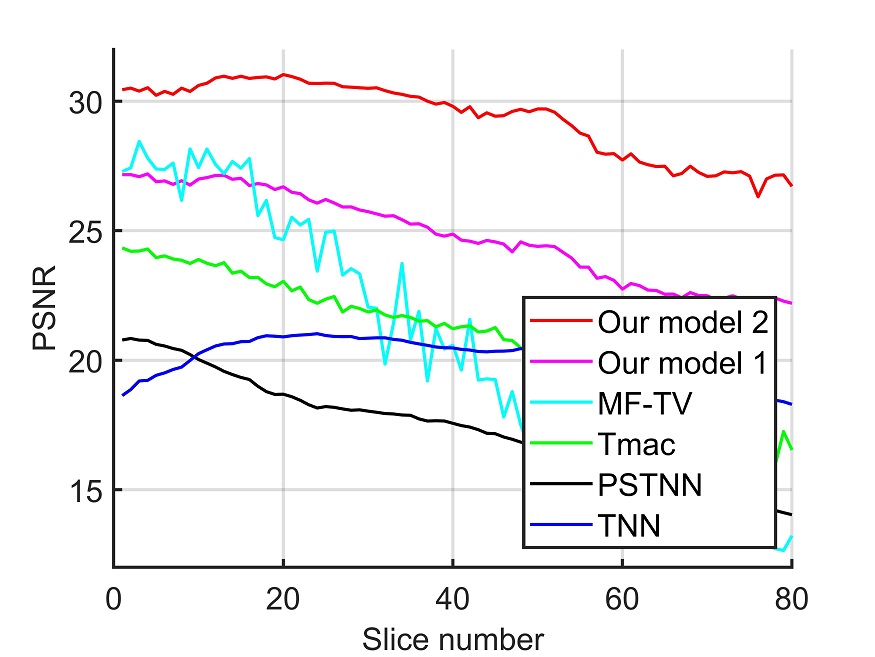}
\hfil
\includegraphics[width=0.3\linewidth]{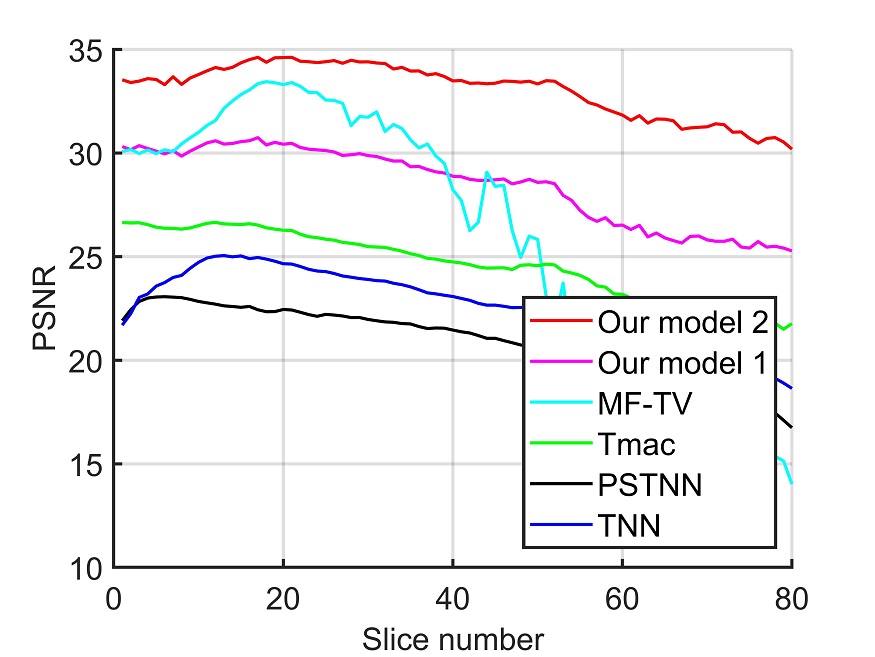}
\hfil
\includegraphics[width=0.3\linewidth]{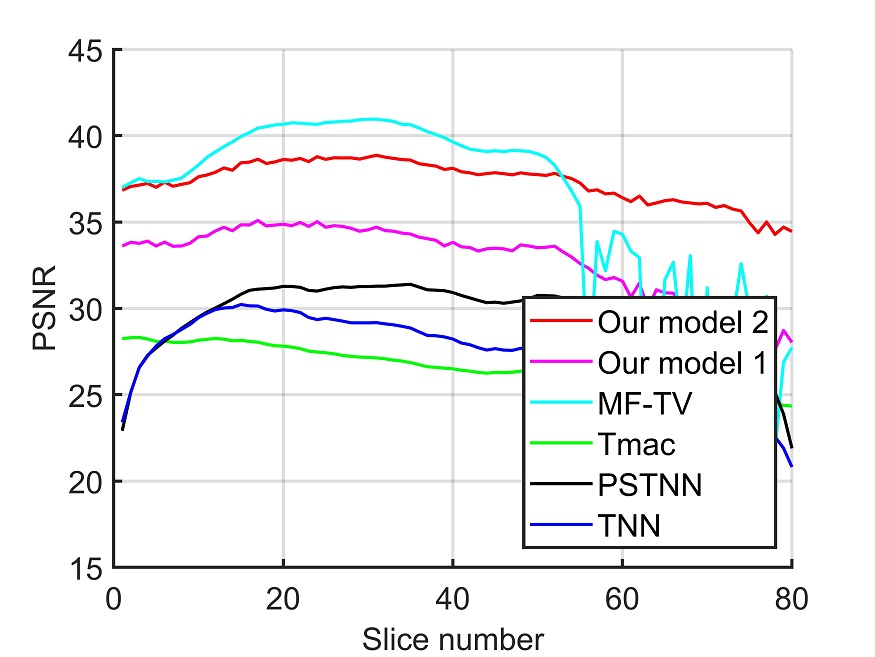}
\hfil
\includegraphics[width=0.3\linewidth]{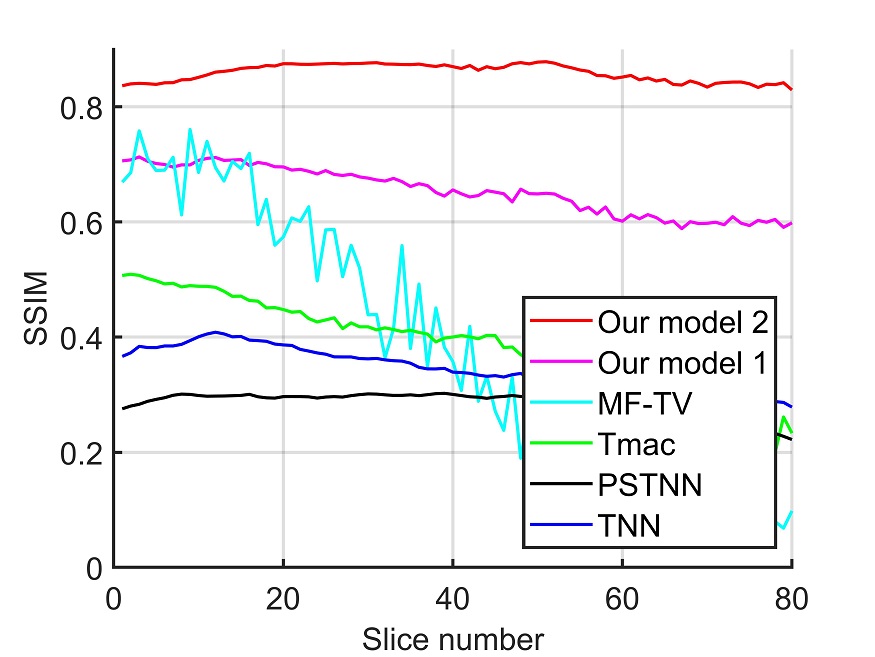}
\hfil
\includegraphics[width=0.3\linewidth]{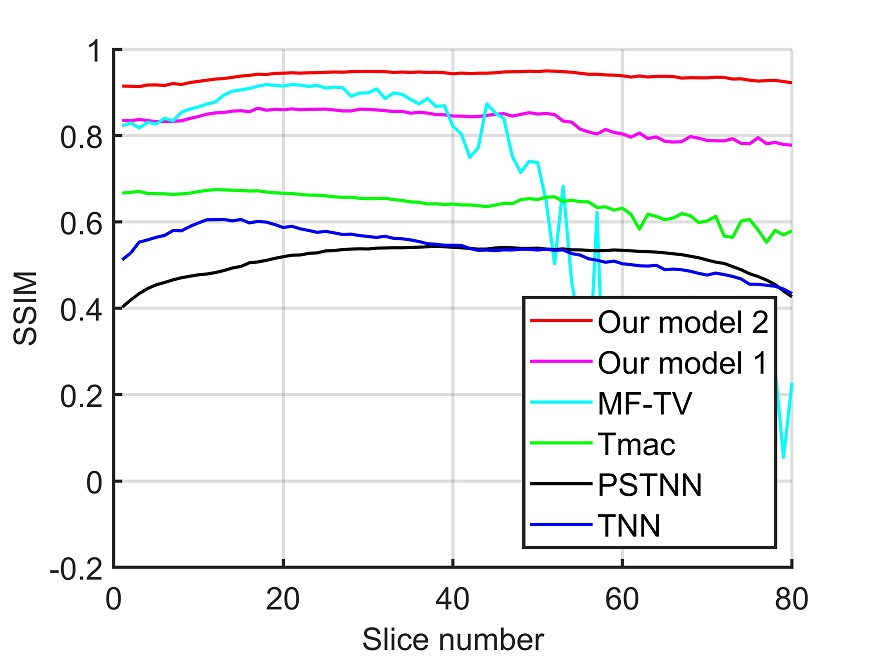}
\hfil
\includegraphics[width=0.3\linewidth]{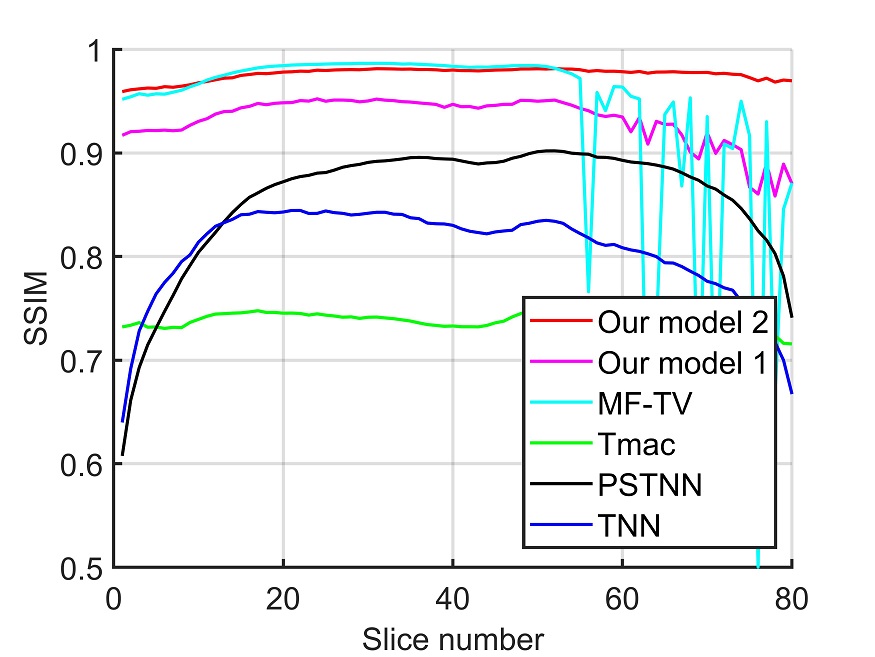}	
\hfil				
\subfloat[SR = 0.025]{\includegraphics[width=0.3\linewidth]{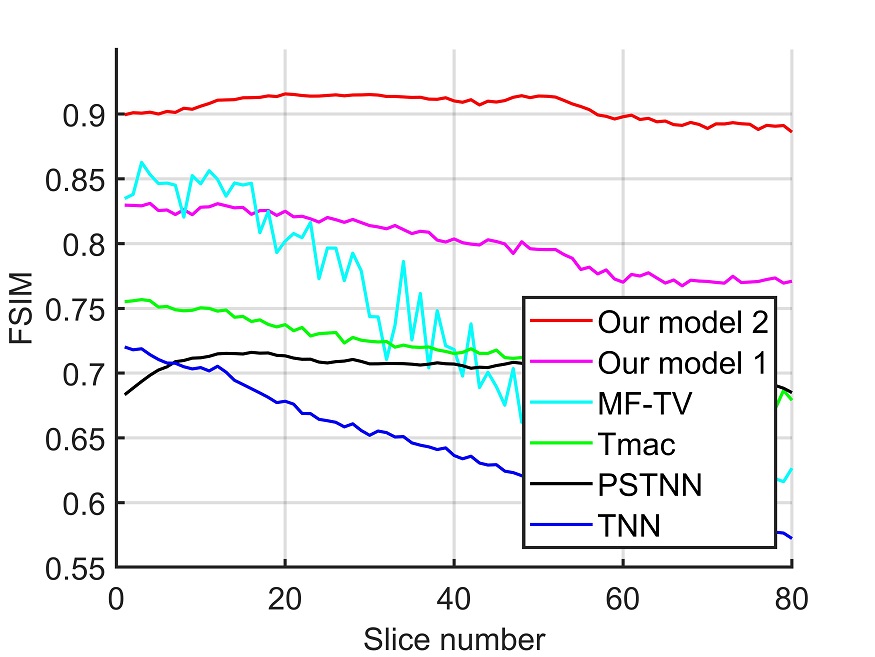}}%
\hfil
\subfloat[SR = 0.05]{\includegraphics[width=0.3\linewidth]{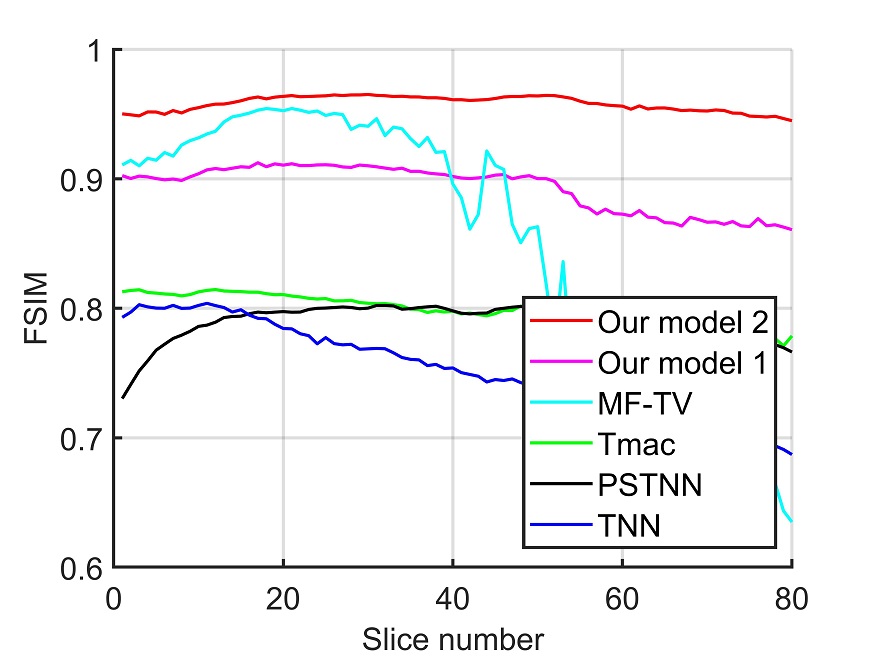}}%
\hfil
\subfloat[SR = 0.1]{\includegraphics[width=0.3\linewidth]{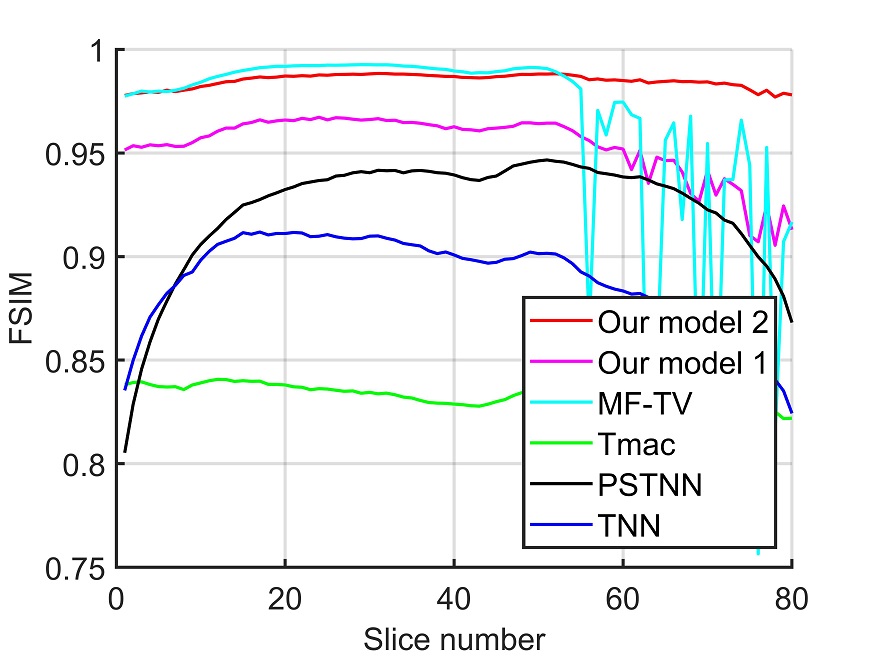}}%
\caption{The PSNR, SSIM and FSIM of the recovered HSI "Pavia" by MF-TV, Tmac, TNN, PSTNN and our model-1 and model-2 for all slices, respectively.}
\label{PSNR and SSIM of Pavia}
\end{figure*}

Table \ref{table_HSI} and Table \ref{table_Pavia} list the PQIs of the results restored by the proposed models and the competition model at three sampling rates.
Fig. \ref{PSNR and SSIM of HSI} lists the PSNR, SSIM and FSIM of each frontal slice of the recovered "Cuprite" for all methods at sampling rates of 0.025 and 0.05.
Fig. \ref{figure_HSI_sr0.05} shows one slice of the recovered "Cuprite" for all methods at sampling rate of 0.05.
Fig. \ref{PSNR and SSIM of Pavia} lists the PSNR, SSIM and FSIM of each frontal slice of the recovered "Pavia" for all methods at sampling rates of 0.025, 0.05 and 0.1. Fig. \ref{figure_Pavia_sr0.025} and Fig. \ref{figure_Pavia_sr0.05} shows one slice of the recovered "Pavia" for all methods at sampling rates of 0.025 and 0.05.
It can be clearly seen that the two proposed methods not only obtain the higher PQIs,
but also recover the more structure information of the image, and restore more spatial details than comparison methods, especially at low sampling rates.
Therefore, one can see that the recovered data obtained by our models has the best visual evaluation and PQIs.

\begin{table}[!t]
\caption{The averaged PSNR, SSIM, FSIM, ERGA and SAM of the recovered results on hyperspectral image "Cuprite" by Tmac, MF-TV, TNN, PSTNN and the proposed model-1, model-2 with different sampling rates. The best value is highlighted in bolder fonts.}
\label{table_HSI}
\resizebox{\textwidth}{42mm}{
	\begin{tabular}{ccccccccc}
		\hline \hline
		&&&SR =0.025&&&&	\\
		method&	Nosiy&	our model-2&	our model-1&	MF-TV&	Tmac&	PSTNN&	TNN\\	
		PSNR	&	7.666	&	\textbf{34.983}	&	31.985	&	26.115	&	21.25	&	13.387	&	22.783	\\
		SSIM	&	0.007	&	\textbf{0.877}	&	0.807	&	0.539	&	0.412	&	0.124	&	0.554	\\
		FSIM	&	0.48	&	\textbf{0.91}	&	0.861	&	0.765	&	0.755	&	0.613	&	0.775	\\
		ERGA	&	1043.633	&	\textbf{47.3}	&	64.636	&	237.074	&	235.594	&	539.574	&	245.333	\\
		MSAM	&	81.221	&	\textbf{1.483}	&	1.833	&	12.913	&	7.842	&	17.98	&	9.156	\\	
		\hline
		&&&SR = 0.05&&&&\\	
		method&	Nosiy&	our model-2&	our model-1&	MF-TV&	Tmac&	PSTNN&	TNN\\		
		PSNR	&	7.779	&	\textbf{38.433}	&	35.402	&	34.684	&	28.945	&	20.621	&	26.579	\\
		SSIM	&	0.01	&	\textbf{0.936}	&	0.893	&	0.845	&	0.712	&	0.31	&	0.663	\\
		FSIM	&	0.471	&	\textbf{0.959}	&	0.928	&	0.915	&	0.846	&	0.735	&	0.836	\\
		ERGA	&	1030.139	&	\textbf{34.53}	&	45.581	&	89.372	&	93.352	&	234.445	&	154.292	\\
		MSAM	&	77.268	&	\textbf{1.225}	&	1.481	&	4.386	&	3.278	&	7.886	&	5.413	\\
		\hline
		&&&SR = 0.1&&&&\\	
		method	&	Nosiy	&	our model-2	&	our model-1	&	MF-TV	&	Tmac	&	PSTNN	&	TNN	\\
		PSNR	&	8.013	&	\textbf{41.182}	&	39.084	&	40.888	&	35.627	&	35.51	&	35.015	\\
		SSIM	&	0.014	&	\textbf{0.961}	&	0.946	&	0.957	&	0.885	&	0.907	&	0.897	\\
		FSIM	&	0.451	&	\textbf{0.979}	&	0.968	&	0.978	&	0.931	&	0.951	&	0.943	\\
		ERGA	&	1002.75	&	\textbf{28.338}	&	33.934	&	34.263	&	44.518	&	54.421	&	57.537	\\
		MSAM	&	71.695	&	\textbf{1.098}	&	1.25	&	1.46	&	1.445	&	2.072	&	2.192	\\
		\hline \hline
\end{tabular}}
\end{table}

\begin{table}[!t]
\caption{The averaged PSNR, SSIM, FSIM, ERGA and SAM of the recovered results on hyperspectral image "Pavia" by Tmac, MF-TV, TNN, PSTNN and the proposed model-1, model-2 with different sampling rates. The best value is highlighted in bolder fonts.}
\label{table_Pavia}
\resizebox{\textwidth}{42mm}{
	\begin{tabular}{ccccccccc}
		\hline \hline
		&&&SR =0.025&&&&	\\
		method&	Nosiy&	our model-2&	our model-1&	MF-TV&	Tmac&	PSTNN&	TNN\\	
		PSNR	&	13.388	&	\textbf{29.353}	&	24.781	&	20.132	&	20.765	&	17.39	&	19.984	\\
		SSIM	&	0.014	&	\textbf{0.859}	&	0.654	&	0.385	&	0.379	&	0.288	&	0.344	\\
		FSIM	&	0.436	&	\textbf{0.905}	&	0.801	&	0.726	&	0.717	&	0.705	&	0.639	\\
		ERGA	&	787.86	&	\textbf{124.945}	&	211.791	&	450.292	&	344.451	&	496.92	&	369.941	\\
		MSAM	&	81.947	&	\textbf{6.612}	&	8.622	&	36.188	&	16.917	&	34.622	&	14.16	\\
		\hline
		&&&SR = 0.05&&&&\\	
		method	&	Nosiy	&	our model-2	&	our model-1	&	MF-TV	&	Tmac	&	PSTNN	&	TNN	\\				
		PSNR	&	13.5	&	\textbf{33.066}	&	28.46	&	25.442	&	24.6	&	21.028	&	22.557	\\
		SSIM	&	0.025	&	\textbf{0.938}	&	0.832	&	0.643	&	0.64	&	0.512	&	0.538	\\
		FSIM	&	0.469	&	\textbf{0.958}	&	0.893	&	0.835	&	0.798	&	0.791	&	0.752	\\
		ERGA	&	777.776	&	\textbf{81.615}	&	139.446	&	316.6	&	216.234	&	326.121	&	276.113	\\
		MSAM	&	77.699	&	\textbf{5.024}	&	7.267	&	26.414	&	10.31	&	22.871	&	13.08	\\
		\hline
		&&&SR = 0.1&&&&\\	
		method	&	Nosiy	&	our model-2	&	our model-1	&	MF-TV	&	Tmac	&	PSTNN	&	TNN	\\	
		PSNR	&	13.736	&	\textbf{37.347}	&	32.818	&	36.064	&	26.523	&	29.518	&	27.363	\\
		SSIM	&	0.046	&	\textbf{0.976}	&	0.932	&	0.938	&	0.737	&	0.857	&	0.806	\\
		FSIM	&	0.516	&	\textbf{0.985}	&	0.954	&	0.965	&	0.832	&	0.923	&	0.889	\\
		ERGA	&	756.907	&	\textbf{50.103}	&	85.931	&	93.891	&	172.184	&	132.084	&	163.529	\\
		MSAM	&	71.908	&	\textbf{3.727}	&	5.896	&	9.397	&	8.871	&	11.236	&	10.825	\\
		\hline \hline
\end{tabular}}
\end{table}

%
%

\section{Conclusions}

In this paper, we propose two new low-rank models based on multiple mode matrix decomposition for tensor completion.
Instead of the traditional single nuclear norm, we adopt a double nuclear norm to represent the low-rank structure in all modes of underlying tensors, and propose our model-1.
Further, in order to preserve the local smoothing structure of the target tensors,
we introduce the total variation regularization into model-1, and propose our model-2.
The BSUM can be used to efficiently solve our models, and it can be demonstrated that our numerical scheme converge to the coordinatewise minimizers.
The proposed models have been evaluated on three types of public datasets,
which show that our algorithms can recover a variety of low-rank tensors with significantly fewer samples than the compared methods.

%
%
%
%
%

\bibliography{mybibfile}

\end{document}